\documentclass[a4paper]{article}

\usepackage{lineno}
\modulolinenumbers[5]
\usepackage{xurl}

\usepackage{graphicx}
\usepackage{amsmath,amsfonts,mathrsfs,amssymb}
\usepackage{bm}
\usepackage[noend]{algpseudocode}
\usepackage{hyperref}
\usepackage{subfig}
\usepackage{color}
\usepackage{xcolor}
\usepackage{rotating}
\usepackage{multirow}
\usepackage{placeins}
\usepackage{soul}
\usepackage{marvosym}
\usepackage{fancyhdr}
\usepackage{cite}
\usepackage{caption}

\newcommand{\sep}{ }

\newcommand{\clusRF}{\textsc{CLUS-RF}}

\newcommand{\bigO}{\mathcal{O}}

\begin{document}
\title{Semi-supervised Predictive Clustering Trees for (Hierarchical) Multi-label Classification}

\author{Jurica Levati\'{c}$^{1,2}$, Michelangelo Ceci$^{3,1}$, Dragi Kocev$^{1,2}$, Sa\v{s}o D\v{z}eroski$^{1,2}$\\
\textit{$^{1}$Jo\v{z}ef Stefan Institute, Ljubljana, Slovenia}\\
\textit{$^{2}$Jo\v{z}ef Stefan International Postgraduate School, Ljubljana, Slovenia}\\
\textit{$^{3}$Department of Computer Science, University of Bari, Italy}\\
\{Jurica.Levatic@ijs.si, Michelangelo.Ceci@uniba.it,\\ Dragi.Kocev@ijs.si, Saso.Dzeroski@ijs.si\}}

\maketitle

\begin{abstract}
Semi-supervised learning (SSL) is a common approach to learning predictive models using not only labeled, but also unlabeled examples. While SSL for the simple tasks of classification and regression has received much attention from the research community, this is not the case for complex prediction tasks with structurally dependent variables, such as multi-label classification and hierarchical multi-label classification. These tasks may require additional information, possibly coming from the underlying distribution in the descriptive space provided by unlabeled examples, to better face the challenging task of simultaneously predicting multiple class labels.
In this paper, we investigate this aspect and propose a (hierarchical) multi-label classification method based on semi-supervised learning of predictive clustering trees, which we also extend towards ensemble learning. Extensive experimental evaluation conducted on 24 datasets shows significant advantages of the proposed method and its extension with respect to their supervised counterparts. Moreover, the method preserves interpretability of classical tree-based models. 

\end{abstract}
Semi-supervised learning, \sep Structured output prediction, \sep Multi-label classification, \sep Hierarchical multi-label classification, \sep Decision trees, \sep Random forests


\section{Introduction}


Over the past decade, there has been growing interest for machine learning methods that can use both labeled and unlabeled data for learning a classification model.
This interest is motivated by two important factors: \textit{i)} the high cost of assigning labels for large datasets and domains where labelling requires complex procedures and/or tedious manual effort; \textit{ii)} the opportunity to achieve greater predictive performance by better estimation of the distribution of data in the descriptive space, given the large amount of freely available unlabeled data.
While the former factor is only of practical relevance, the latter stems from the theoretical observation that the underlying marginal data distribution $p(X)$ over the descriptive space $X$ might contain information about the posterior distribution $p(Y|X)$ for the prediction of the  $Y$ values in the target space. 
The machine learning setting that takes into account both motivating factors is semi-supervised learning (SSL) \cite{chapelle_semi-supervised_2006}. It accommodates the second factor by leveraging three (not independent) theoretical assumptions \cite{DBLP:journals/ml/EngelenH20}: the \textit{smoothness} assumption (if two samples $x$ and $x'$ are close in the input space, their labels $y$ and $y'$ should be the same); the \textit{low-density} separation assumption (the decision boundary should not cut through high-density areas of the input space); and the \textit{manifold} assumption (data points on the same low-dimensional manifold should have the same label).

Nowadays, many semi-supervised learning approaches are available that tackle the classification in multiple domains, including object recognition in images \cite{DBLP:conf/nips/JeongLKK19}, 
human speech recognition \cite{YU2010433}, protein 3D structure prediction \cite{DBLP:journals/eswa/LevaticCSDK20}, IoT data analysis \cite{10.1145/3620677}, and spam filtering \cite{10.1145/1835449.1835598}.
However, only a few approaches are suited for the more complex tasks of multi-label classification (MLC) or hierarchical multi-label classification (HMLC), even though many applications (including the ones listed above) have an inherent complexity suitable for MLC and HMLC.
Multi-label classification is a predictive machine learning task where the examples can be labeled with more than one of the labels from a predefined set of labels $C$. In this case, the output variable $y$ takes values in a subset of the label set $\mathcal{Y}$ (i.e., $y \in 2^C$).
Hierarchical multi-label classification is a particular case of MLC, where the output space is structured so that it accommodates dependencies between labels. In particular, labels are organized in a hierarchy: An example labeled with label $c$ is also labeled with all parent/super-labels of $c$.
MLC and HMLC problems are encountered in various domains, such as text categorization, image classification, object/scene classification, gene function prediction, prediction of compound toxicity, etc \cite{Kocev13:jrnl}.
A common property for MLC and HMLC domains is that obtaining labeled examples is harder and more expensive compared to the classical (i.e., single-label) classification context. This contributes greatly to the need for developing SSL methods tailored for the MLC and HMLC tasks.

In the literature, only a few existing approaches tackle the problem of semi-supervised multi-label classification and hierarchical multi-label classification. Some examples include the work presented in \cite{DBLP:conf/kdd/ShiSLG20,DBLP:conf/pkdd/GuoS12,DBLP:journals/tkde/KongNZ13} for the task of SSL MLC and that presented in \cite{DBLP:journals/eswa/SantosC14} for SSL HMLC. However, all these methods adopt generative or optimization-based approaches, yielding complex and time-demanding learning processes, which produce non-interpretable models. On the contrary, in this paper, \emph{we propose an approach to SSL MLC/HMLC based on predictive clustering trees} (PCTs) \cite{blockeel_top-down_proc_1998}.
The advantage of predictive clustering trees is manifold: 
1) The learning phase is time efficient;
2) The SSL models are interpretable for both MLC and HMLC tasks;
3) The SSL models can take both quantitative and categorical variables into account;
4) PCTs can combined into ensembles, such as random forests, to further improve their predictive performance;
5) The hierarchical structure of tree-based models can naturally model the hierarchical structure of the output space in the HMLC task.


In this paper, we propose a method for MLC and HMLC that works in the SSL setting. It defines a novel algorithm for learning predictive clustering trees by exploiting both the labeled and unlabeled data for MLC and HMLC tasks. In a nutshell, this is achieved by defining a new heuristic and prototype functions that take these specifics into account. Moreover, the proposed method has a parameter that balances the contribution of the descriptive and the target/label part of the data (i.e., controlling the degree of supervision in the model learning process). This mechanism safeguards against performance degradation, compared to learning only from the labeled data. Furthermore, we propose learning ensembles of the semi-supervised predictive clustering trees to further boost their predictive performance. The extensive experiments across 24 datasets from a variety of domains reveal that the proposed methods have better predictive performance compared to their supervised counterparts.

\begin{figure}[tb]
		\includegraphics[height=0.44\textwidth]{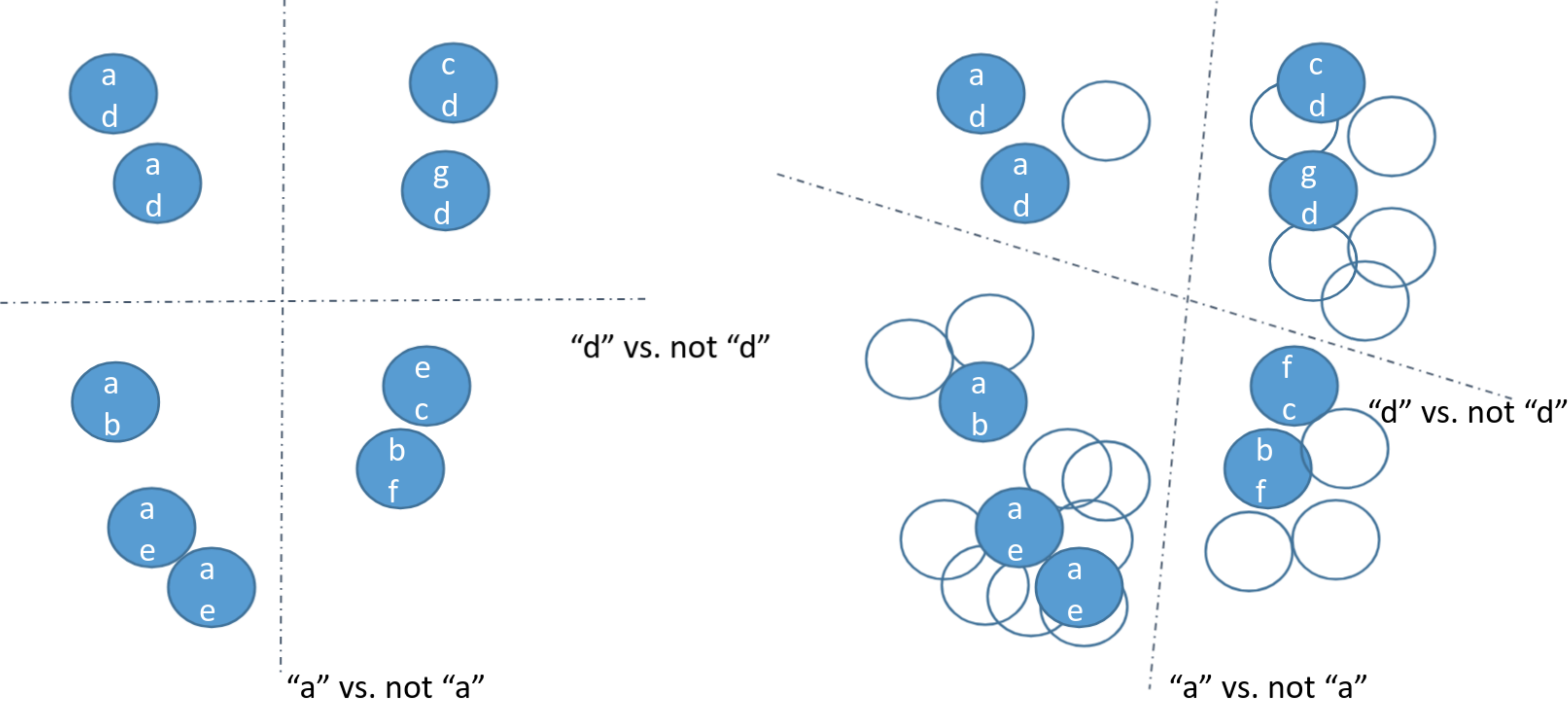}
 \begin{tabular}{lcr}
 a) labeled examples only \ \ \ \ \ \ & \ \ \ \ \ & \ \ b) labeled and unlabeled examples
 \end{tabular}
	\caption{\label{fig:motivating example} Semi-supervised learning in multi-label classification. Filled circles represent labeled examples, while empty circles represent unlabeled examples. Letters represent class labels. }
\end{figure}

One of the explanations of the inner workings of the proposed methods is related to the interaction of semi-supervised learning with the label dependency of MLC and HMLC tasks. More specifically, we investigate whether the \textit{smoothness} assumption (and, indirectly, since they are not independent, the \textit{low-density} and the \textit{manifold} assumptions) holds in the MLC and HMLC contexts. Intuitively, better identification of the distribution of examples in the descriptive space (as performed in the semi-supervised learning setting) can lead to better exploitation of label dependency and/or label correlation in the output space, leading to improved predictive accuracy. 
To better explain this concept, let us consider the example reported in Figure \ref{fig:motivating example}. We can see that unlabeled examples provide useful information to better classify the examples in the classes "a" ("a" vs. not "a", see the vertical dashed lines) and "d" ("d" vs. not "d", see the horizontal dashed lines), especially in low-density regions. From Figure \ref{fig:motivating example}b we can also see that unlabeled examples reveal a higher correlation between the class labels "a" and "e" than between "a" and other class labels, such as "d". This is because "a" and "e" appear together in a region that is much denser than the region where "a" and "d" appear together. Such information, if exploited by the predictive model, can be used to better classify MLC examples.

In summary, the main contributions of this paper are as follows:
\begin{itemize}
    \item Novel semi-supervised methods based on predictive clustering trees and random forest ensembles able to deal with both MLC and HMLC tasks.
    \item A semi-supervised method able to produce interpretable MLC and HMLC models.
    \item A mechanism safeguarding the proposed method from the degradation of predictive performance.  
    \item An extensive evaluation and analysis of the proposed method across 12 MLC and 12 HMLC datasets.
\end{itemize}

The rest of the paper is structured as follows: in Section \ref{sec:relatedwork}, we briefly describe the work in the literature that is related to the present paper; in Section \ref{sec:method}, we describe the proposed solution, while, in Section \ref{sec:experiments}, we evaluate its performance on publicly available datasets and discuss the results. Finally, in Section \ref{sec:conclusions}, we present the conclusions of this work and outline possible directions for future research.

\section{Related Work and Motivations}
\label{sec:relatedwork}

SSL MLC is a relatively recent topic in machine learning and data mining. 
One of the most prominent works in this research area is \cite{DBLP:conf/pkdd/GuoS12}, where the main idea is 
to combine large-margin MLC with unsupervised subspace learning. This is done by jointly solving two problems: 1) learning a subspace representation of the labeled and unlabeled inputs and 2) learning a large-margin supervised multi-label classifier on the labeled part of the data. The proposed algorithm works in a single optimization step, which results in a high-time complexity process. To alleviate the problem, the authors proposed a learning procedure which is based on subgradient search and coordinate descent.

In \cite{DBLP:journals/tkde/KongNZ13} the authors propose an SSL MLC algorithm based on the optimization problem of estimating label concept compositions (label co-occurrence). Specifically, the algorithm derives a closed-form solution to this optimization problem and then assigns label sets to the unlabeled instances in a transductive setting.

In \cite{DBLP:conf/kdd/ShiSLG20} the authors propose a deep generative model to describe the label generation process for the SSL MLC task. 
For this purpose, the generative model incorporates latent variables to
describe the labeled/unlabeled data as well as the labeling process. 
A sequential inference model is then used to approximate the model posterior and infer the ground truth labels. The same inference model is then used to predict the label of unlabeled instances.


More recently, in \cite{DBLP:conf/aaai/WangLQS020} the authors proposed a dual Relation Semi-supervised Multi-label Learning (DRML) approach which
jointly explores the feature distribution and the label relation
simultaneously. In this paper, a dual-classiﬁer domain adaptation strategy
is proposed to exploit both the feature distribution and the label relation between examples. Therefore, the optimization simultaneously takes into account instance-level relations across labeled and unlabeled samples in feature space and the relations across labels. This approach has been only applied in the image multi-label classification task.

In \cite{DBLP:conf/ijcai/ZhaoG15} the authors address the task of multi-label learning with incomplete labels, by combining the label imputation function and multi-label prediction function in a mutually beneficial manner. Specifically, the proposed method conducts automatic label imputation within a low-rank and sparse matrix recovery framework while simultaneously performing vector-valued multi-label learning and exploiting unlabeled data with vector-valued manifold regularization. 

The semi-supervised multi-label learning task has also been investigated in the context of graph-structured data by incorporating the idea of label embedding to capture both network topology and higher-order multi-label correlations \cite{DBLP:conf/cikm/SongMZK21}. In this work, the label embedding is generated along with the node embedding based on the topological structure to serve as the prototype center for each class. Moreover, the similarity of the label embedding and node embedding is used as a confidence vector to guide the label smoothing process, obtained by margin ranking optimization to learn the second-order relations between labels.

In \cite{DingScott} the authors derive an extension of the Manifold
Regularization algorithm to multi-label classification in graph data. They then augment the algorithm with a weighting strategy to allow differential influence on a model between instances having ground truth vs. induced labels. Therefore, the proposed approach includes three components: the graph construction, the manifold regularization with multiple labels, and the exploitation of a reliance weighting strategy.


All the previously mentioned works, although they tackle the 
SSL MLC problem, suffer from the common problem of not generating interpretable models. 
This is not the case with the method proposed in this paper, where the adoption of the PCT framework allows us to produce multi-label decision trees, which are directly interpretable and fast to learn (a preprint of this work has previously been published \cite{levatic2022semi}).
Moreover, contrary to existing approaches, the approach we propose builds models by exploiting clustering. This allows us to take into account the smoothness assumption, 
both for the descriptive space and for the output space. 
Finally, the mentioned existing approaches cannot be directly used to impose limitations on the labels, and, therefore, cannot be directly used for the more complex task of HMLC.

As for the SSL HMLC task, the existing work in the literature is relatively limited.
In \cite{DBLP:journals/eswa/SantosC14} the authors extend the RA$k$EL system, initially developed for (supervised) MLC, to the SSL HMLC task, leading to three new methods, called HMC-SSBR, HMC-SSLP and HMC-SSRA$k$EL. 
RA$k$EL is an ensemble-based wrapper method for solving MLC tasks using existing algorithms for multi-class classification. The idea is to build the ensemble by providing a small random subset of $k$ labels (organized as a label powerset) to each base model, learned by a multi-class classifier. This approach is also used in HMC-SSBR, HMC-SSLP and HMC-SSRA$k$EL, which, therefore, are not based on clustering and cannot directly take into account the smoothness, the low-density, and the manifold assumptions.

In the more general context of semi-supervised structured output prediction, some approaches for \textit{multi-target regression} also use predictive clustering trees. This is the case of the works in \cite{DBLP:journals/kbs/LevaticCKD17} and \cite{DBLP:journals/isci/LevaticKCD18}, where the idea is to learn predictive clustering trees by using both labeled and unlabeled examples. \cite{DBLP:journals/kbs/LevaticCKD17} proposed a semi-supervised multi-target regression method based on the self-training approach with a random forest of predictive clustering trees. In self-training a model is trained iteratively with its own most reliable predictions. \cite{DBLP:journals/isci/LevaticKCD18} 
extended multi-target regression PCTs by adapting the heuristics used for the construction of the trees, in order to consider both labeled and unlabeled examples. Both methods, however, do not tackle the classification tasks.

\section{Background: Predictive clustering trees}

The predictive clustering trees (PCTs), presented in this paper for MLC and HMLC, are inspired by the work in \cite{blockeel_top-down_proc_1998}. In that work, the splits in the tree are evaluated by considering both descriptive and target attributes. The semi-supervised decision trees proposed here have similarities to the ones in \cite{blockeel_top-down_proc_1998}, with multiple differences. First, \cite{blockeel_top-down_proc_1998} considered unlabeled examples only in tasks with primitive outputs, whereas we designed semi-supervised trees for structured outputs. Second, we established a parameter that allows varying degrees of supervision in the trees (i.e., how much the descriptive attributes influence the evaluation of the splits). In this way we can build supervised, semi-supervised, or unsupervised trees, dictated by the demands of the specific dataset we are dealing with. 

The PCT framework\footnote{PCTs are implemented in the CLUS system \cite{PETKOVIC2023101526} available for download at \url{https://github.com/knowledge-technologies/clus}.} treats a decision tree as a hierarchically organized set of clusters, where the topmost cluster contains all the data. This cluster is recursively divided into smaller clusters as one moves from the root to the leaves, generating PCTs. PCTs represent a generalization of default decision trees (e.g., C4.5 \cite{Quinlan93:book}) where the outputs are more complex structures than in conventional classification and regression tasks. Classical PCTs can predict several types of structured outputs, including nominal/real value tuples, class hierarchies, and short time series \cite{Kocev13:jrnl}. For each type, two functions must be defined: the prototype function and the variance function. The prototype function associates a class label to each leaf in the tree and it returns a representative structured value (i.e., a prototype). The variance function evaluates the homogeneity of a set of such structured values and is used to find the best splits while constructing the tree.  

In this study, we propose semi-supervised PCTs and ensembles of semi-supervised PCTs, for the tasks of MLC and HMLC. Thus, in Sections~\ref{sec:pct_mlc} and \ref{sec:pct_hmlc}, we present \emph{supervised} PCTs for these tasks in more detail.

To build an ensemble model for predicting structured output, an appropriate type of PCTs is utilized as a base model. For example, to build an ensemble for the HMLC task, PCTs for HMLC are used as base models. An ensemble predicts a new example by considering predictions of all the ensemble's base models. For regression tasks, predictions of the base models are averaged, while for classification tasks, various strategies can be used, such as the probability-based majority voting, which we used as suggested by \cite{Bauer99:jrnl}. According to this strategy, each base tree provides the probability of an example belonging to each of the possible classes. The class with the highest sum of probabilities, considering all of the base trees, is predicted.

\subsection{PCTs for multi-label classification}\label{sec:pct_mlc}

The variance function for learning PCTs for the MLC task computes the average of the $Gini$ indices across all the target variables. For a set of examples $E$ with target space $Y$, consisting of $T$ nominal target variables $Y_1, Y_2, \ldots, Y_T$, the variance function is defined as follows: 
\begin{equation}
Var_f(E, Y) = \frac{1}{T} \cdot \sum_{i=1}^T Gini(E,Y_i),
\label{eq:variance_mlc_all}
\end{equation}
where $Gini(E,Y_i)$ is the $Gini$ score of the $i^{th}$ target variable $Y_i$ for a set of examples $E$. The Gini score of the $i^{th}$ target variable is calculated as follows:
\begin{equation}
Gini(E,Y_i) = 1 - \sum_{j=1}^{C_i}p_j^2,
\end{equation}
where $C_i$ is the number of classes for the target variable $Y_i$ (e.g., if $Y_i$ is binary, then $C_i=2$), and $p_j$ is the \emph{apriori} empirical probability of a class $c_j$ (i.e., the relative frequency of instances in $E$ that belong to the class $c_j$).

The sum of the entropies of class variables can also be used as a variance indicator, i.e., $Var_f(E,Y) = \sum_{i=1}^T \mathit{Entropy(E,Y_i)}$ (this was considered previously for MLC \cite{Clare03:phd}). The CLUS framework includes other variance functions as well, such as reduced error, gain ratio and the $m$-estimate.

The prototype function returns a vector denoting probabilities of an instance belonging to a given class for each target variable. To determine the predicted classes, the user can specify a threshold on probabilities, or the majority class (i.e., the most probable one) for each target can be calculated. In this study, we use the majority class.

\subsection{PCTs for Hierarchical multi-label classification}\label{sec:pct_hmlc}

In HMLC, the target space $Y$ is associated to a hierarchy of classes ($C,\leq_h$), where $\forall {c_l, c_j \in C} : c_l \leq_h c_j \iff c_l \text{~is a superclass of~} c_j$. The set of labels of example $e_i$ is represented as a binary vector $L_i$, whose $j^{th}$ component is 1 if the example is labeled with the class $c_j$, 0 otherwise. The $j^{th}$ component of the arithmetic mean of such vectors contains the relative frequency of examples of the set belonging to class $c_j$. Then, the variance indicator over a set of examples $E$ is calculated as the average squared distance between each vector $(L_{i})$ and the set's mean class vector $(\overline{L})$: 
\begin{equation}
Var_f(E, Y) = \frac{1}{|E|} \cdot \sum\limits_{e_{i} \in E} d(L_{i},\overline{L})^2.
\label{eq:variance_hmc_all}
\end{equation}

In the HMLC context, the similarities at higher levels of the hierarchy are considered to be more important than the similarities at lower levels. The distance measure in the above formula (weighted Euclidean distance) is therefore defined as follows:
\begin{equation}
\label{eq:w_eucl}
d(L_1,L_2) = \sqrt{\sum\limits_{l = 1}^{|C|} \omega(c_l)\cdot(L_{1,l}-L_{2,l}})^2,
\end{equation}
where $L_{i,l}$ is the $l^{th}$ component of the class vector $L_i$ of an instance $e_i$, $|C|$ is the number of classes in the hierarchy (i.e., the size of the class vector), and the class weights $\omega(c)$ decrease with the depth of the class in the hierarchy.
More precisely, $\omega(c) = \omega_0^{depth(c)}$, where $depth(c)$ denotes the depth of the class $c$ in the hierarchy, and $0 < w_0 < 1$. Note that class weights can be calculated recursively, i.e., $\omega(c) = \omega_0 \cdot \omega(par(c))$, where $par(c)$ denotes the parent of class $c$. In this work we use $\omega_0 = 0.75$, as recommended by \cite{Vens08:jrnl}.

The definition of $\omega(c)$ is general enough to represent classes that are organized as a directed acyclic graph (DAG). Generally, a DAG-like hierarchy can be interpreted in two ways: an example belonging to a class $c$, either \textit{i)} belongs to all super-classes of $c$, or \textit{ii)} belongs to one or more superclasses of $c$. In this work, we consider the former.

The variance indicator for tree-structured hierarchies uses the weighted Euclidean distance between the class vectors (as defined in Eq.~\ref{eq:w_eucl}), where the weight of a class changes depending on its level in the hierarchy. Note that in DAG-shaped hierarchies, the classes do not have a unique level number. To resolve this issue, we follow the recommendation of \cite{Vens08:jrnl}: The weight of a given class is calculated as an average of all the weights according to possible paths from the root to that class.    

In classification trees, a leaf holds the majority class of its examples, which the tree predicts for examples arriving in that leaf. In the HMLC task, an example can have multiple classes, so the meaning {\it majority class} is not straightforward. The prediction, in this case, is a mean $\bar{L}$ of the class vectors of the examples in the leaf. The $i^{th}$ component of the vector $\bar{L}$ can be considered as the probability that an example in the leaf belongs to class $c_i$.
The final classification for an example that arrives in the leaf can be made using a threshold $\tau$ for the probabilities; if $\bar{L_i}>\tau$, then class $c_i$ is predicted for the example. When making predictions, the parent-child relationships from the class hierarchy are preserved if the values for the thresholds $\tau$ are defined as follows: $\tau_i \leq \tau_j$ whenever $c_i \leq_{h} c_j$ ($c_i$ is an ancestor of $c_j$). The selection of the threshold $\tau$ depends on the use scenario, e.g., trading off higher precision with lower recall. Here, we use a threshold-independent metric based on precision-recall curves to evaluate the predictive performance of the models.

\section{Semi-supervised PCT learning for MLC and HMLC}
\label{sec:method}

\subsection{Task definition}

Here, we formally define the semi-supervised learning tasks for the types of structured outputs considered in this study: predicting multiple targets and hierarchical multi-label classification.
 
\paragraph{Semi-Supervised Multi-label classification} In MLC, the task is to predict several binary values (i.e., labels) for each example. More formally:
 
 \textbf{Given:}
 \begin{itemize}
 	\item A descriptive (or input) space $X = X_1 \times X_2 \times \ldots \times X_D$ spanned by $D$ descriptive variables that consist of values of primitive data types (Boolean, nominal or continuous).
 	
 	\item A target (or output) space $Y = Y_1 \times Y_2 \times \ldots \times Y_T$ spanned by the $T$ binary target variables.
 	
 	\item A set of labeled examples $E_l = \left\{(x_i, y_i) \mid x_i \in X, y_i \in Y, 1 \leq i \leq N_l\right\}$, where each example $e_i \in E_l$ is described according to both the descriptive space and the target space, and $N_l$ is the number of labeled examples.
 	
 	\item A set of unlabeled examples $E_u = \left\{(x_i) \mid x_i \in X, 1 \leq i \leq N_u\right\}$, where each example $e_i \in E_u$ takes its values from the descriptive space only, and $N_u$ denotes the number of unlabeled examples.
 	
 	\item A quality metric $q$, e.g., which favours models with high predictive accuracy (or low predictive error).
 \end{itemize}
 
 \textbf{Find:} A function $f : X \rightarrow Y$ that maximizes $q$.

 \paragraph{Semi-Supervised Hierarchical Multi-label classification} In HMLC, each example can have more than one class (multiple labels), and the classes are organized in a hierarchical structure, i.e., an example belonging to a class also belongs to all its superclasses. More formally:
 
 \textbf{Given:}
 \begin{itemize}
 	\item A descriptive (or input) space $X = X_1 \times X_2 \times \ldots \times X_D$ spanned by $D$ descriptive variables that consist of values of primitive data types (Boolean, nominal or continuous).
 	
 	\item A target space $Y$, defined with a hierarchy of classes ($C,\leq_h$), where $C$ is a set of classes and $\leq_h$ is a partial order among them, representing the superclass relationship, i.e., $\forall c_1, c_2 \in C, c_1 \leq_h c_2$ if and only if $c_1$ is a superclass of $c_2$.
 	
 	\item A set of labeled examples $E_l = \left\lbrace \left(x_i,Y_i\right) \mid x_i \in X, Y_i \subseteq C, 1 \leq i \leq N_l\right\rbrace$, where each example $e_i \in E_l$ is a pair of a tuple $x_i$ from the descriptive space and a set $S_i$ from the target space, and each set satisfies the hierarchy constraint, i.e., $c \in S_i, c' \in C, c' \leq_h c \Rightarrow c' \in Y_i$, and $N_l$ is the number of labeled examples. 
 	 
 	 \item A set of unlabeled examples $E_u = \left\{(x_i) \mid x_i \in X, 1 \leq i \leq N_u\right\}$, where each example $e_i \in E_u$ takes its values from the descriptive space only, and $N_u$ is the number of unlabeled examples.
 	
 	\item A quality metric $q$, e.g., which favours models with high predictive accuracy (or low predictive error).
 \end{itemize}
 
 \textbf{Find:} a function $f : X \rightarrow 2^C$ (where $2^C$ is the power set of $C$) such that $f$ maximizes $q$ and the predictions made by $f$ satisfy the hierarchy constraint, i.e., $c \in f(x), c' \in 2^C, c' \leq_h c \Rightarrow c' \in f(x)$.

\subsection{Tree learning}

The proposed semi-supervised algorithm (see Table~\ref{alg:clus}) is based on the extension of the standard {\it top-down induction of decision trees} (TDIDT) algorithm used to build supervised PCTs \cite{Breiman84:book}. An input to the TDIDT algorithm is a set of examples $E$. The heuristic ($h$) selects the best tests ($t^*$) based on the reduction of the variance resulting from partitioning ($\mathcal{P}$) the examples (BestTest function in Table~\ref{alg:clus}). As the variance reduction is maximized, the homogeneity of the cluster is also maximized. If no suitable test is found, i.e., if none of the candidate tests results in a significant reduction of the variance or if there are fewer examples in a node than the specified limit, then a leaf is created and the prototype of the examples in that leaf is computed.

\begin{table}[!h]
		\centering
		\caption{\label{alg:clus} The proposed algorithm for learning of semi-supervised predictive clustering trees.}
		\begin{footnotesize}
			\begin{tabular} {cc}
				\hline
				\begin{minipage}[t]{0.4\textwidth}
					{\bf procedure} SSL-PCT\\
					{\bf Input:} A dataset $E = E_l \cup E_u$, a $w$ parameter\\
					{\bf Output:} A predictive clustering tree
					\begin{algorithmic}[1]
						\State $(t^*,h^*,\mathcal{P}^*) = \mathrm{BestTest}(E, w)$
						\If{$t^* \neq \mathit{none}$}
						\For{\textbf{each} $E_i \in \mathcal{P^*}$}
						\State $\mathit{tree}_i$ = SSL-PCT($E_i$, $w$)
						\EndFor
						\State \textbf{return} $\mathrm{node}(t^*,\;\bigcup_i \{\mathit{tree}_i\})$
						\Else
						\State \textbf{return} $\mathrm{leaf}(\mathrm{Prototype}(E))$ 
						\EndIf
					\end{algorithmic}
				\end{minipage} &
				\begin{minipage}[t]{0.5\textwidth}
					{\bf procedure} $\mathrm{BestTest}$\\
					{\bf Input:} A dataset $E$, a $w$ parameter\\
					{\bf Output:} The best test ($t^{*}$), its heuristic score ($h^{*}$) and the partition ($\mathcal{P}^{*}$) it induces on the dataset ($E$)
					\begin{algorithmic}[1]
						\State $(t^*,h^*,\mathcal{P}^*) = (\mathit{none},0,\emptyset)$
						\For{\textbf{each} possible test $t$} \label{alg:btest1}
						\State $\mathcal{P} = $ partition induced by $t$ on $E$
						\State $h = \mathit{Var_f}(E,Y,X,w) - \sum_{E_i \in \mathcal{P}} \frac{|E_i|}{|E|} \mathit{Var_f}(E_i,Y,X,w)$ \label{alg:icv}
						\If{$(h > h^*) \land \mathrm{Acceptable}(t,\mathcal{P})$}
						\State $(t^*,h^*,\mathcal{P}^*) = (t,h,\mathcal{P})$ \label{alg:btest2}
						\EndIf
						\EndFor
						\State \textbf{return} $(t^*,h^*,\mathcal{P}^*)$
					\end{algorithmic}
				\end{minipage}\\
			\end{tabular}
               \begin{tabular} {cc}
                \begin{minipage}[t]{0.4\textwidth}
                \vspace{0.1cm}
                {\bf procedure} OptimizeParamW\\
                {\bf Input:} A dataset $E = E_l+E_u$; a set of values $W$, $\forall w \in W, w \in [0, 1]$; a number of folds k\\
                {\bf Output:} A $w$ value
                \begin{algorithmic}[1]
                \For{\textbf{each}  $w\in W$}
                \State $w_{opt}$ = \newline $\underset{w \in W}{{arg\,max}}\ {CrossValidate}(E, w, k)$ 
                
                \State \textbf{return} $w_{opt}$
                \EndFor
                \end{algorithmic}
                \end{minipage} &
                \begin{minipage}[t]{0.5\textwidth}
                \vspace{0.1cm}
                {\bf procedure} CrossValidate\\
                {\bf Input:} A dataset $E = E_l \cup E_u$, $w \in [0,1]$, a number of folds {k}\\
                {\bf Output:} An accuracy measure
                \begin{algorithmic}[1]
                \State $E_l^{\{1,2,\ldots, k\}}$ = partition $E_l$ into $k$ folds
                \For{\textbf{each} j $\in$ \{1,2,\ldots, k\}}
                \State $\mathit{tree}_j$ = SSL-PCT($E_l^{\{1,2,\ldots, k\} \backslash j } \cup E_{u}$, w)
                \State $accuracy_j$ = evaluate($\mathit{tree}_j$, $E_l^j$)
                \EndFor
                \State \textbf{return} $\sum_{j=1}^k{\frac{accuracy_j}{k}}$
                \end{algorithmic}
                \end{minipage}\\
                \hline
            \end{tabular}
            \\[5pt]
            \caption*{As an input, the SSL-PCT algorithm uses a set of labeled examples ($E_l$), a set of unlabeled examples ($E_u$), and a $w \in [0,1]$ parameter, The $w$ parameter is optimized using the procedure that relies on internal cross-validation.}
		\end{footnotesize}
\end{table}

The supervised TDIDT algorithm for PCTs is extended towards semi-super\-vised learning as follows. First, the input to the SSL algorithm dataset comprises both labeled and unlabeled examples: $E=E_l \cup E_u$, where $E_{l}$ are labeled examples and $E_{u}$ are unlabeled examples. Second, the variance function in the SSL algorithm considers both the target and the descriptive attributes in the evaluation of splits. It is calculated as a weighted sum of the variance over the target space $Y$ and the variance over the descriptive space $X$:
\begin{equation}
Var_f(E,Y,X,w) = w \cdot Var_f(E,Y) + (1-w) \cdot Var_f(E,X),
\label{eq:variance_SSL_all}
\end{equation}
where $w \in \left[0,1\right]$ is the parameter that controls the trade-off between the contribution of the target space and the descriptive space to the variance function. During the learning of semi-supervised regression trees, the $w$ parameter is automatically optimized by an internal cross-validation procedure (OptimizeParamW function in Table~\ref{alg:clus}).  

This extension relies on the semi-supervised cluster assumption \cite{chapelle_semi-supervised_2006}: \textit{If examples are in the same cluster, then they are likely to be of the same class}. We recall that, the variance function of supervised PCTs uses only the target attributes (Eqs. \ref{eq:variance_mlc_all} and \ref{eq:variance_hmc_all}). Consequently, (a) unlabeled examples cannot contribute to the tree construction (since only their descriptive attributes are known), and (b) the clusters produced by supervised PCTs are only homogeneous regarding the class label. Enforcing the similarity of examples in both the descriptive and the target space during the construction of SSL-PCTs results in clusters that are homogeneous regarding both the descriptive and the target space. This allows us to exploit both labeled and unlabeled examples. Finally, following the cluster assumption, labeled and unlabeled examples that end up in the same leaf of the tree are likely to be of the same class.   

Parameter $w$ controls the magnitude of the contribution that unlabeled examples have on the learning of semi-supervised PCTs. In other words, parameter $w$ enables learned models to range from fully supervised ($w=1$) to completely unsupervised ($w=0$). The control of the contribution of unlabeled examples enabled by the $w$ parameter allows us to set the amount of supervision for different datasets appropriately. This aspect is discussed in more detail in the experimental analysis (Section~\ref{sec:wparam_sslpct_mtr}).

The variance of a set of examples $E$ on \textit{target} space $Y$ is calculated differently, depending on the type of structured output at hand:
\begin{equation}\label{eq:var_ssl_target}
Var_f(E,Y) =
\begin{cases}
\frac{\sum_{i=1}^T Gini(E,Y_i)}{T}, \text{ if } Y \text{ consists of } T \text{ binary variables} \\[10pt]
\frac{\frac{1}{|E|} \cdot \sum\limits_{e_{i} \in E_l} d(L_{i},\overline{L})^2}{\sum\limits_{l = 1}^{|C|} w(c_l)}, \text{ if } Y \text{ is a hierarchy of classes}
\end{cases}
\end{equation}

Since the descriptive variables can be either numeric or nominal, the variance on the descriptive space of a set of examples $E$ is computed as follows:
\begin{equation}
Var_f(E,X) = \frac{1}{D} \cdot \left( \sum_{X_i \text{ is numeric}}Var(E,X_i) + \sum_{X_j \text{ is nominal}}Gini(E,X_j) \right),
\label{eq:variance_SSL_all_desc}
\end{equation}
where $D$ is the number of descriptive attributes and 
the variance or the Gini score of descriptive attributes is calculated following Eqs. ~\ref{eq:var_ssl_numatt} and \ref{eq:var_ssl_nomatt}.

Let $N$ be the number of examples (both labeled and unlabeled), and let $K_i$ be the number of examples with non-missing values of the $i^{th}$ attribute $Y_i$. Then, the variance for the continuous attributes and the Gini index for the nominal attributes are calculated as follows, respectively:
\begin{equation}
\label{eq:var_ssl_numatt}
Var(E,Y_i) = \frac{ \frac{N-1}{K_i-1} \cdot \sum_{j=1}^{K_i} (y_{i,j})^2 - N \cdot \left(\frac{1}{K_i} \cdot \sum_{j=1}^{K_i} y_{i,j} \right)^2}{N},
\end{equation}
\begin{equation}
\label{eq:var_ssl_nomatt}
Gini(E,Y_i) = 1 - \sum_{j=1}^{C_i}{\left(\frac{\left|\lbrace{e : e \in E \wedge c_j \in classes(e) \rbrace}\right|}{K_i}\right)}^2 = 1 - \sum_{j=1}^{C_i}\hat{p_j},
\end{equation} 
where $C_i$ is the number of class values of $Y_i$, and $\hat{p_j}$ is the \textit{apriori} probability of class value $c_j$, estimated by using only examples for which the value for variable $Y_i$ is known. Note that for the HMLC task, the variance for the output space is calculated only on the labeled data (see Eq.~\ref{eq:var_ssl_target}).

The variances of descriptive and target attributes are normalized, similarly to supervised PCTs, to ensure the equal contribution of attributes to the final variance. Normalization is performed by dividing the variance estimates of individual attributes in Eq.~\ref{eq:var_ssl_numatt} and Eq.~\ref{eq:var_ssl_nomatt} (that consider the set of examples in the current node of the tree) with the variance of the corresponding attribute considering the entire training set.


During the semi-supervised tree construction phase, two extreme cases can occur: (1) Only unlabeled examples can end up in a leaf of the tree, therefore, the prototype function cannot be calculated, or (2) variance needs to be calculated for attributes where none of the examples (or only one) have non-missing values (e.g., $K_i \leq 1$ in Eq.~\ref{eq:var_ssl_numatt}). For the first extreme case, we calculate the prototype function of such a leaf by returning the prototype of its first parent node that contains labeled examples. In other words, from a leaf that contains only unlabeled examples, we move up the tree until we encounter a node containing labeled examples and we return the prototype of such a node. The prototype is calculated using only labeled examples as described in Sections~\ref{sec:pct_mlc} and \ref{sec:pct_hmlc}. Nodes with only unlabeled examples are not split further, while in leaf nodes containing labeled examples, we allow a minimum of 2 labeled examples. Both criteria can be considered as ``stopping criteria'', to stop the tree construction phase. Note that, these criteria are coherent with the stopping criteria implemented in supervised PCTs, where at least two labeled examples in a leaf node are required.

The second extreme case can occur when the examples in a node are split in a way that only unlabeled examples go into a single branch of the tree. In such a case, a split needs to be evaluated with one of the branches containing only missing values for the target attribute(s); therefore, variance for such attributes cannot be calculated. Similarly as in the first extreme, we handle this situation by using the variance of the parent node (for the attributes containing only missing values in the original node). Note that, since we do not split nodes with only unlabeled examples, the parent node is guaranteed to contain labeled examples. 

\subsubsection{Semi-supervised PCTs with Feature weighting}

PCTs (and decision trees in general) are robust to irrelevant features since the learning algorithm chooses only the most informative features when building (supervised) trees. Thus, irrelevant features will be ignored. However, in semi-supervised PCTs this feature may be compromised, since the evaluation of the splits depends on both target and descriptive attributes. To deal with this issue, we propose feature-weighted \textsc{SSL-PCT}s.

Methods for feature weighting can be used to identify the most informative features by determining an importance score (weight), where a higher score denotes more informative features, while a lower score denotes less informative ones. The effectiveness of feature weighting with the importance scores was shown to help the k-nearest neighbors algorithm to deal with irrelevant features \cite{cunningham_k-nearest_2007}. Similarly, we adapt the \textsc{SSL-PCT}s and use importance scores to assign weights to features.

More specifically, we use a feature ranking method based on a random forest of PCTs \cite{Kocev13:jrnl}, to obtain the importance score $\hat{\sigma_i}$ for each descriptive attribute $X_i$. To calculate feature importance, this method uses the internal out-of-bag (OOB) error as an estimate of the noise in the descriptive space. The rationale is that, if noise is introduced to a descriptive variable which is important, then the error of the model will increase (as measured by OOB error estimates).

The feature ranking is performed on the labeled examples $E_l$ prior to building \textsc{SSL-PCT}s or \textsc{SSL-RF}s. 
The importance scores are then normalized as follows: $\sigma_i = \hat{\sigma_i} / \max (\hat{\sigma_1}, \hat{\sigma_2}, \ldots, \hat{\sigma_D})$. The function for the calculation of the variance of the descriptive attributes of \textsc{SSL-PCTs} is then adapted to include normalized feature importance scores $\sigma_i$ as weights of the descriptive attribute $X_i$:
\begin{equation}
\begin{split}
Var_f(E,X) = \frac{1}{D} \cdot \sum_{X_i \text{ is numeric}} \sigma_i \cdot Var(E,X_i) + \\ \frac{1}{D} \cdot \sum_{X_j \text{ is nominal}} \sigma_j \cdot Gini(E,X_j),
\label{eq:variance_SSL_all_desc_FR}
\end{split}
\end{equation}

This results in irrelevant features contributing less to the variance score. Henceforth, semi-supervised PCTs and random forests with feature weighting are denoted as \textsc{SSL-PCT-FR} and \textsc{SSL-RF-FR}, respectively.

\subsection{Semi-supervised random forests}

\textsc{SSL-PCTs} can be easily extended to their random forest version \cite{Breiman01a:jrnl}. This is done by using \textsc{SSL-PCT}s as the members of the random forest ensemble, instead of using classical supervised trees. The notable difference is, however, in the presence of both labeled and unlabeled examples in the bootstrap samples, which does not conform with the classical random forest algorithm \cite{Breiman01a:jrnl}. Namely, the trees can be built with only a small set of labeled examples and a large set of unlabeled examples, thus bootstrap samples may end up containing only unlabeled examples. In order to overcome this problem, in the semi-supervised setting, we perform stratified bootstrap sampling where the proportions of labeled and unlabeled examples are preserved in each bootstrap sample. For example, if the training data contains 10\% of labeled and 90\% of unlabeled examples, such a ratio is maintained in bootstrap samples. This is achieved by separately sampling labeled and unlabeled examples and later joining them to form a bootstrap sample for the random forest algorithm. 

\subsection{Computational complexity}
	
To assess the complexity of the algorithm for learning SSL-PCTs, we first introduce the computational complexity of learning \textit{supervised} PCTs: sorting of $D$ descriptive variables ($\bigO(DN \log N)$), used to determine the best split for $T$ target variables ($\bigO(TDN)$), for $N$ labeled training examples ($O(N)$). If we assume that the expected depth of the tree is $\bigO(\log N)$ \cite{witten2005data}, the computational complexity of building a single PCT is $\bigO(DN \log^2 N) + \bigO(TND \log N) + \bigO(N \log N)$.

Now we discuss the changes introduced in \textsc{SSL-PCT}s. First, the number of training examples $N$ in the semi-supervised case equals the combined number of unlabeled and labeled examples (i.e., $N=N_l+N_u$, instead of $N=N_l$). Second, SSL-PCTs use both $D$ descriptive variables and $T$ target variables to determine the best split, therefore, this step has the complexity of $\bigO((T+D)ND)$. Therefore, the computational complexity of building an SSL-PCT tree is $\bigO(DN \log^2 N)) + \bigO((D+T)ND \log N) + \bigO(N \log N)$.

The computational complexity of random forests of semi-supervised \textsc{PCT}s is bounded by $k (\bigO(D^{\prime} N^{\prime} \log^{2}N^{\prime}) + \bigO((T+D) N^{\prime} D^{\prime} \log N^{\prime}))$, where $N^{\prime}$ is the number of bootstrap samples, $D^{\prime}$ is the number of features considered at each tree node and $k$ is the number of trees. The added computational complexity of feature ranking is that of randomly permuting the values of the out-of-bag samples ($N^{\prime\prime}=N-N^{\prime}$) and sorting the samples through the tree. Both operations are done for each descriptive attribute and their cost is $O(DN^{\prime\prime} + D\log N)$. This added computational cost is, however, negligible compared to the overall cost of building the random forest ensemble. Note that, the number of examples in feature ranking is $E_l$ because the feature weights are calculated considering only the labeled examples.

\section{Experimental design}
\label{sec:experiments}
In this section, we first describe the datasets used in the experimental evaluation. Next, we present the evaluation procedure, the specific parameter settings of the algorithms and the performance measures.

\subsection{Data description}

To evaluate the proposed methods, we use 24 datasets of the two structured output prediction tasks considered: MLC and HMLC. The datasets are from various domains and have different sizes and numbers of descriptive and target variables. The characteristics of the datasets are summarized in Tables~\ref{table:datasets_sslpct_mlc} and \ref{table:datasets_sslpct_hmlc} for the MLC and HMLC tasks, respectively.

\begin{table}[!tb]
	\begin{center}
            \begin{footnotesize}
    		\caption{\label{table:datasets_sslpct_mlc}MLC datasets and their characteristics.}
    		\begin{tabular}{l l r r r r}			
    			\hline
    			Dataset & Domain & {\bf $N$} & {\bf $D/C$} & {\bf $\mathcal{L}$} & $\overline{\mathcal{L}_L}$\\
    			\hline
    			{\bf Bibtex} \cite{katakis2008multilabel} & Text & 7395 & 1836/0 & 159 & 2.402 \\
    			{\bf Birds} \cite{briggs20139th} & Audio & 645 & 2/258 & 19 & 1.014 \\
    			{\bf Emotions} \cite{trohidis2008multi} & Music & 594 & 0/72 & 6 & 1.869 \\
    			{\bf Corel5k} \cite{Duygulu2002} & Images & 5000 & 499/0 & 374 & 3.522 \\
    			{\bf Enron} \cite{Klimt04:proc} & Text & 1702 & 1001/0 & 53 & 3.378 \\
    			{\bf Genbase} \cite{diplaris2005protein} & Text & 662 & 1186/0 & 27 & 1.252 \\
    			{\bf Mediana} \cite{Skrjanc01:jrnl} & Media & 7953 & 21/58 & 5 & 1.205 \\
    			{\bf Medical} \cite{read2011classifier} & Text & 978 & 1449/0 & 45 & 1.245 \\
    			{\bf Scene} \cite{boutell_learning_2004} & Images & 2407 & 0/294 & 6 & 1.074 \\
    			{\bf SIGMEA real} \cite{Demsar05-SIGMEA:proc} & Ecology & 817 & 0/4 & 2 & 0.726 \\
    			{\bf Slovenian rivers} \cite{Dzeroski00:jrnl} & Ecology & 1060 & 0/16 & 14 & 5.073 \\
    			{\bf Yeast} \cite{elisseeff_kernel_2001} & Biology & 2417 & 0/103 & 14 & 4.237 \\		
    			\hline
    		\end{tabular}
            \\[5pt]
            \caption*{$N$ is the number of examples, $D/C$ is the number of descriptive variables (nominal/continuous), $\mathcal{L}$ is the number of labels and $\overline{\mathcal{L}_L}$ is the average number of labels per example.}
            \end{footnotesize}
        \end{center}
\end{table}

\begin{table}[!tb]
        \begin{footnotesize}
	\begin{center}
		\caption{\label{table:datasets_sslpct_hmlc}HMLC datasets and their characteristics.}
		\begin{tabular}{l l r r c r r r}
			\hline
			Dataset & Domain & {\bf $N$} & {\bf $D/C$} & $\mathcal{H}$ & {\bf $|\mathcal{H}|$} & {\bf $\mathcal{H}_d$} & $\overline{\mathcal{L}_L}$\\
			\hline
			{\bf Danish farms} \cite{Demsar06:jrnl} & Ecology & 1944 & 132/5 & Tree & 72 & 3 & 7 \\
			{\bf Enron} \cite{Klimt04:proc} & Text & 1702 & 1001/0 & Tree & 53 & 2 & 3.38 \\
			{\bf Slovenian rivers} \cite{Dzeroski00:jrnl} & Ecology & 1060 & 0/16 & Tree & 724 & 4 & 25 \\
			{\bf ImCLEF07A} \cite{dimitrovski2011hierarchical} & Images & 11006 & 0/80 & Tree & 96 & 3 & 1 \\
			{\bf ImCLEF07D} \cite{dimitrovski2011hierarchical} & Images & 11006 & 0/80 & Tree & 46 & 3 & 1 \\
                {\bf Diatoms} \cite{dimitrovski2012hierarchical} & Images & 3119 & 0/371 & Tree & 377 & 3 & 0.94 \\
			{\bf Cellcycle-GO} \cite{Vens08:jrnl} & Genomics & 3766 & 0/77 & DAG & 4126 & 12 & 35.91 \\
			{\bf Church-GO} \cite{Vens08:jrnl} & Genomics & 3764 & 1/26 & DAG & 4126 & 12 & 35.89 \\
			{\bf Derisi-GO} \cite{Vens08:jrnl} & Genomics & 3733 & 0/63 & DAG & 4120 & 12 & 35.99 \\
			{\bf Eisen-GO} \cite{Vens08:jrnl} & Genomics & 2425 & 0/79 & DAG & 3574 & 11.12 & 39.04 \\
			{\bf Expr-GO} \cite{Vens08:jrnl} & Genomics & 3788 & 4/547 & DAG & 4132 & 12 & 35.87 \\
			{\bf Pheno-GO} \cite{Vens08:jrnl} & Genomics & 1592 & 69/0 & DAG & 3128 & 12 & 36.43 \\						
			\hline
		\end{tabular}
        \\[5pt]
        \caption*{$N$ is the number of examples, $D/C$ is the number of descriptive variables (nominal/continuous), $\mathcal{H}$ is the type of the label hierarchy, $|\mathcal{H}|$ is the number of nodes in the hierarchy, $\mathcal{H}_d$ is the maximal depth of the hierarchy, $\overline{\mathcal{L}_L}$ is the average number of labels per example.}
	\end{center}
        \end{footnotesize}
\end{table}

\subsection{Experimental setup}

We introduce semi-supervised PCTs (\textsc{SSL-PCT}) and their feature-weighted variant (\textsc{SSL-PCT-FR}). We compare these methods across different structured output prediction tasks with supervised PCT algorithms for MLC and HMLC, denoted as \textsc{SL-PCT}, in order to estimate the contribution of unlabeled data to the predictive performance of the methods under the same conditions. By such comparison, we can answer our main question: \textit{Are SSL-PCTs able to outperform supervised PCTs?}. In the experiments with single trees, we use the pruning procedure as implemented in M5 regression trees \cite{Quinlan93:book}. 

We also compare the predictive performance of semi-supervised random forests (\textsc{SSL-RF}) and their feature-weighted variant (\textsc{SSL-RF-FR}) to supervised random forests for structured output prediction (\clusRF). We use 100 unpruned trees to construct random forests. The number of randomly selected features at each node we set to $ \left \lfloor{\log_2(D) + 1}\right \rfloor $, where $D$ is the total number of features \cite{Breiman01a:jrnl}.


To assess the influence of different proportions of labeled/unlabeled data for the semi-supervised method, we vary the number of labeled examples across the following set of values: \{50, 100, 200, 350, 500\}. 
The labeled examples are randomly sampled from the training set, while the rest of the examples are used both as unlabeled examples and as testing data. We temporarily ignore their labels and use them in the semi-supervised methods as unlabeled training samples. The test set used to evaluate the models comprises the same examples and their original labels restored. The evaluation scenario is thus in the context of transductive learning. The supervised methods are trained using the selected labeled samples and evaluated on the same test set as semi-supervised methods. This is repeated 10 times using different random initialization, while the predictive performances are averaged over the 10 runs.

For each of the 10 runs, we optimize the parameter $w$ (weight) by an internal 3-fold cross-validation procedure performed on the labeled portion of the training set. The semi-supervised methods also use the available unlabeled examples. The values of the parameter $w$ vary from 0 to 1 with a step of 0.1.

The algorithms are evaluated by means of the area under the Precision-Recall curve (AUPRC). Since the considered tasks are MLC and HMLC, we use a variant of the AUPRC -- the area under the micro-averaged average Precision-Recall curve ($\mathrm{AU}\overline{\mathrm{PRC}}$), as suggested by \cite{Vens08:jrnl}. 
Specifically, the precision and recall values are computed as follows:
$$\overline{Prec} = \frac{\sum_i{TP}_i}{\sum_i{TP}_i + \sum_i{FP}_i}, \quad \text{and} \quad \overline{Rec} = \frac{\sum_i{TP}_i}{\sum_i{TP}_i + \sum_i{FN}_i},$$
where $i$ ranges over all the classes.

We statistically analyze the results following the recommendations of \cite{demvsar_statistical_2006}. We use the non-parametric Wilcoxon paired signed-rank test \cite{wilcoxon1945individual} for the comparison of the predictive performance of the two methods over multiple datasets. We set the significance level to 0.05 in all the experiments.

\section{Results and discussion}

\subsection{Predictive performance of the methods}

\subsubsection{Multi-label classification}

\begin{figure}[p]
	\subfloat[Bibtex\label{bibtex_pruned}]{%
		\includegraphics[height=0.3\textwidth]{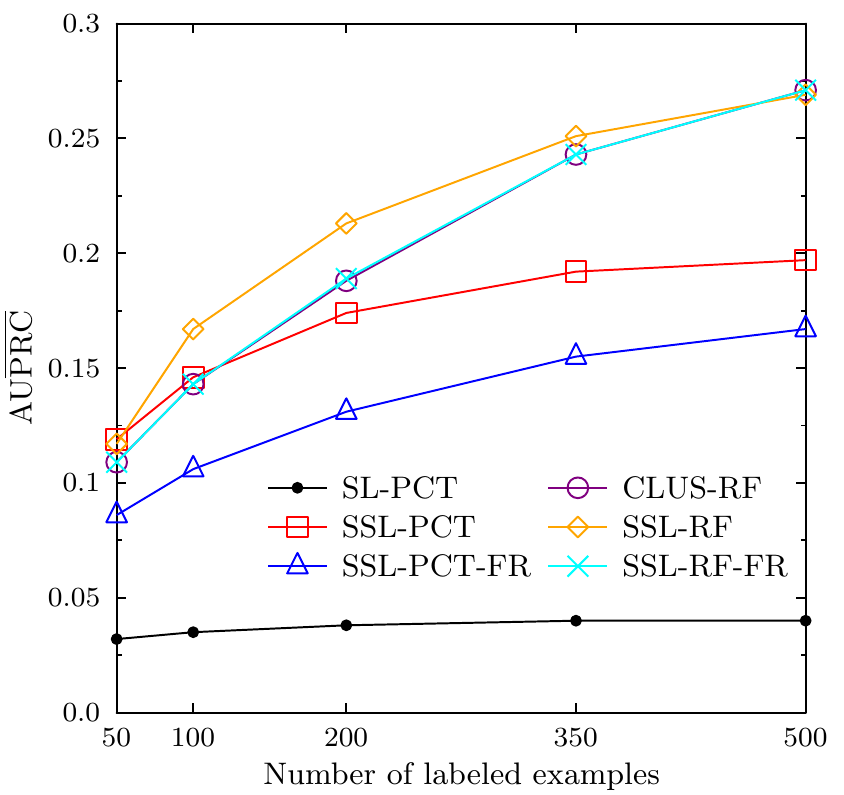}
	}
	\hfill
	\subfloat[Birds\label{birds_pruned}]{%
		\includegraphics[height=0.3\textwidth]{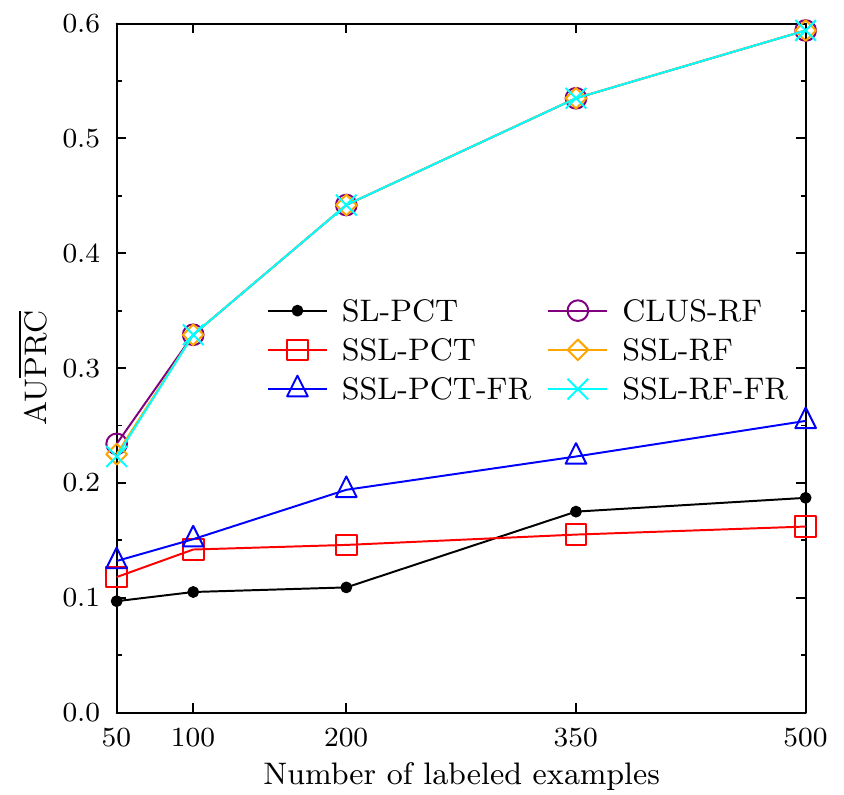}
	}
	\hfill
	\subfloat[Corel5k\label{corel_pruned}]{%
		\includegraphics[height=0.3\textwidth]{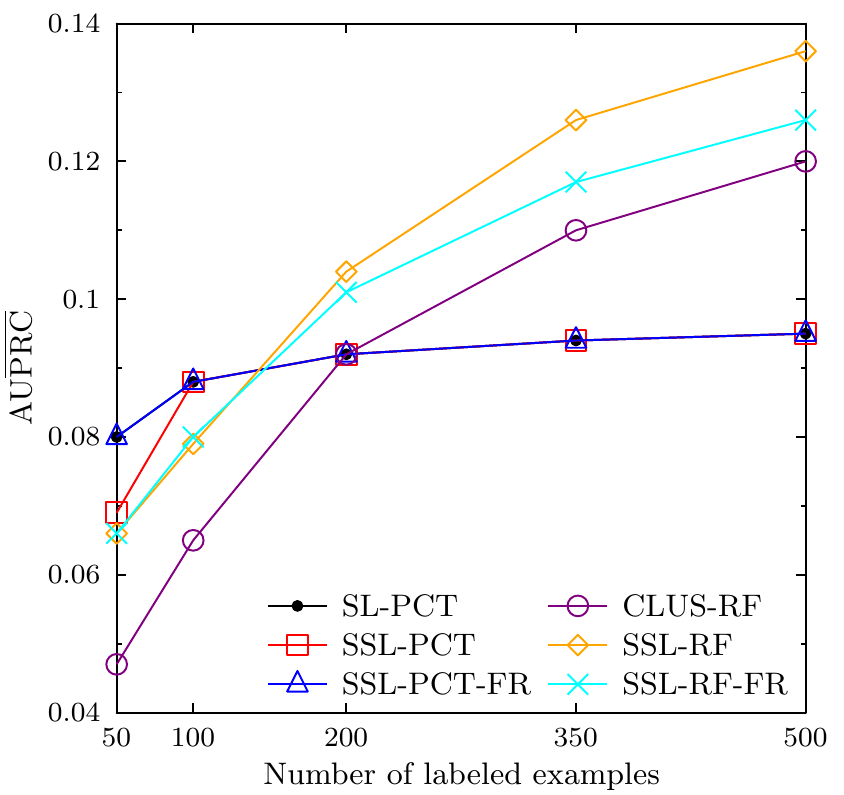}
	}
	\hfill
	\newline
	\subfloat[Emotions\label{emotions_pruned}]{%
		\includegraphics[height=0.3\textwidth]{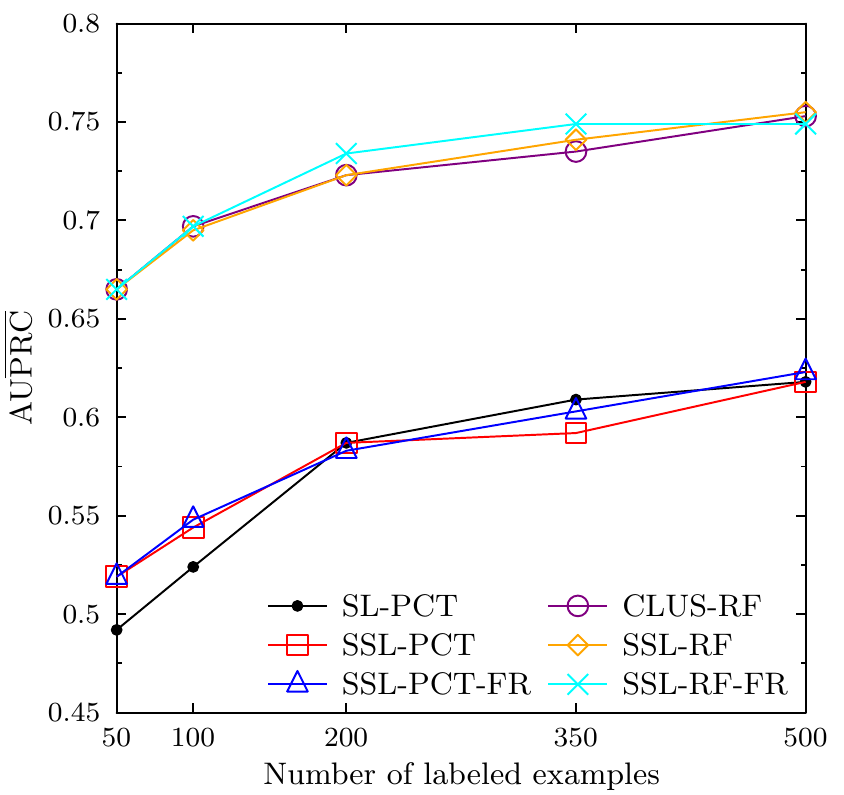}
	}
	\hfill
	\subfloat[Enron\label{enron_pruned}]{%
		\includegraphics[height=0.3\textwidth]{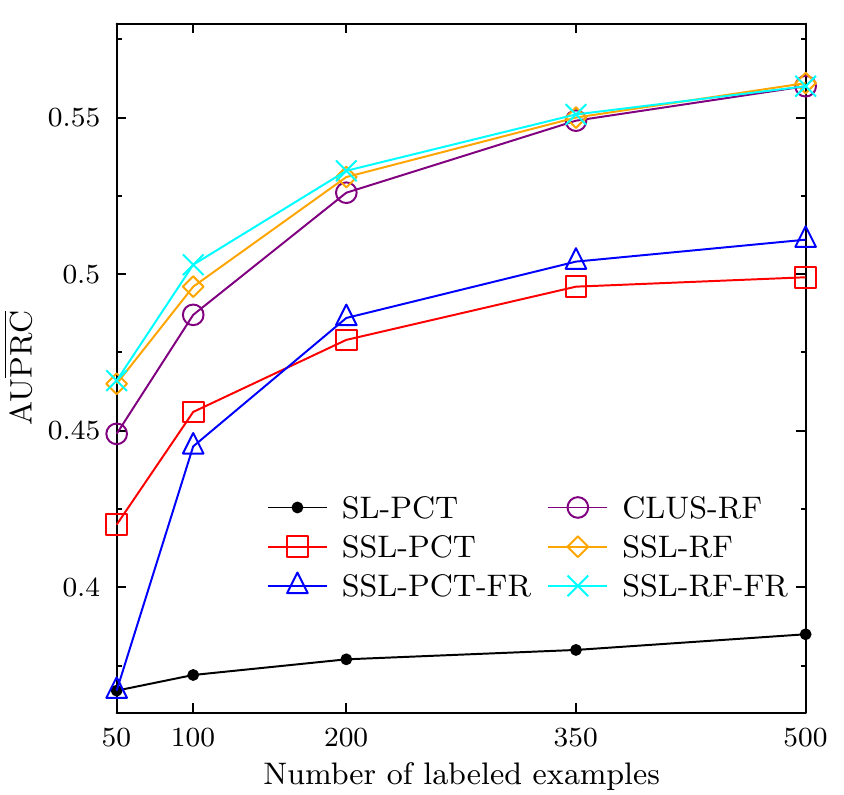}
	}
	\hfill
	\subfloat[Genbase\label{genbase_pruned}]{%
		\includegraphics[height=0.3\textwidth]{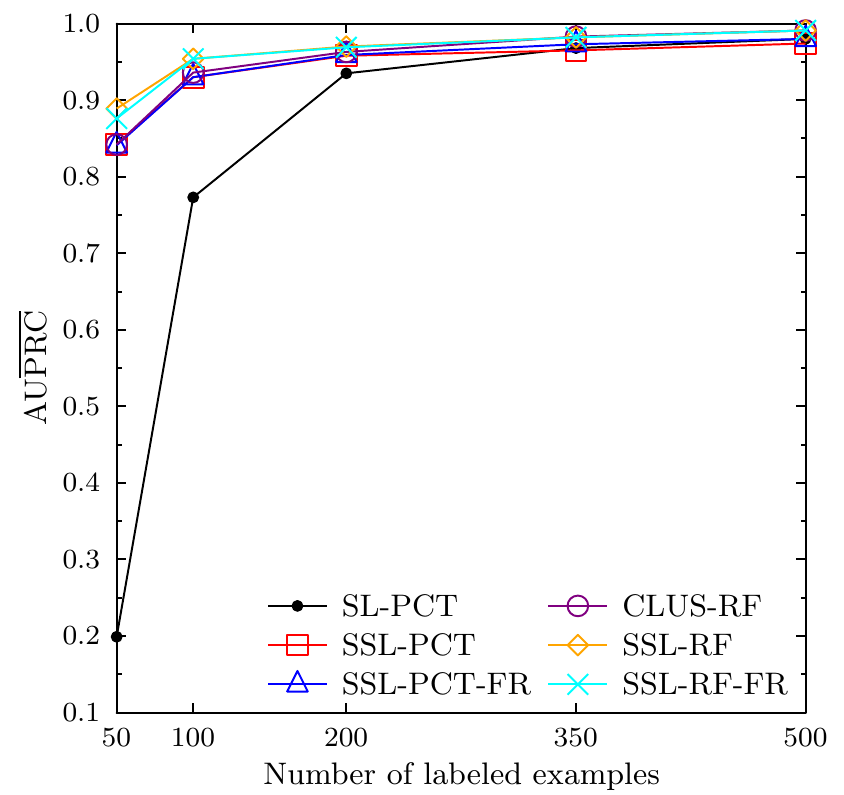}
	}
	\newline
	\subfloat[Mediana\label{mediana_pruned}]{%
		\includegraphics[height=0.3\textwidth]{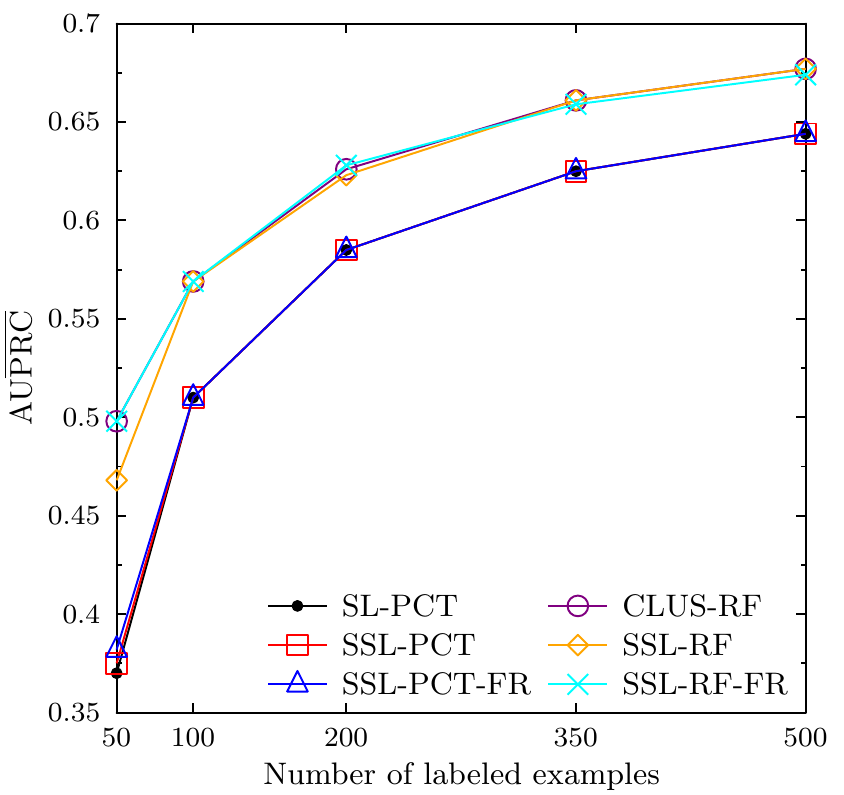}
	}
	\hfill
	\subfloat[Medical\label{medical_pruned}]{%
		\includegraphics[height=0.3\textwidth]{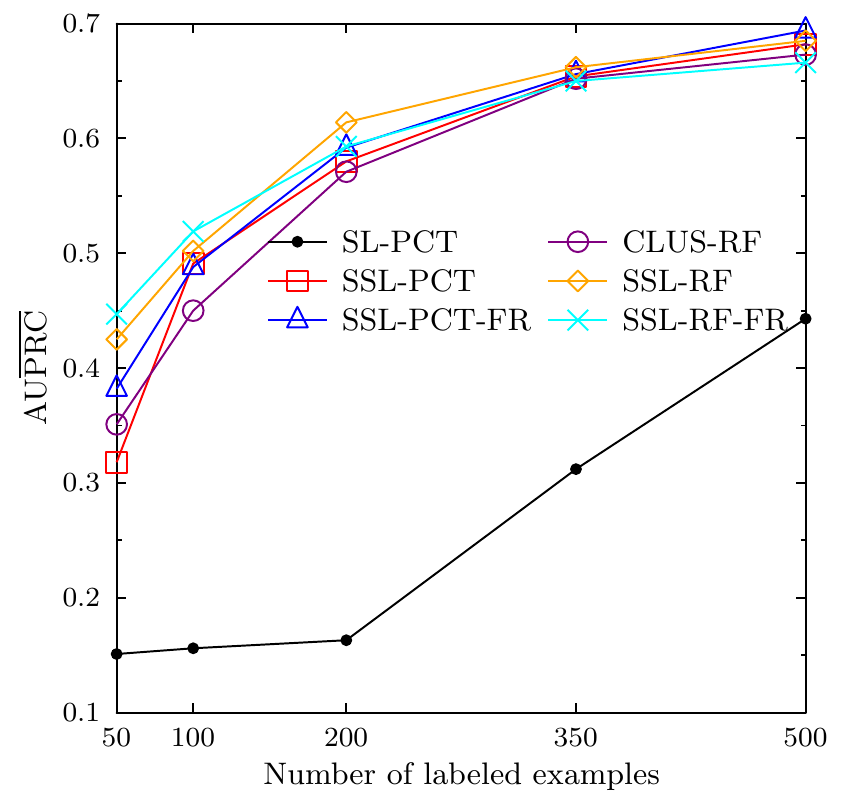}
	}
	\hfill
	\subfloat[Scene\label{scene_pruned}]{%
		\includegraphics[height=0.3\textwidth]{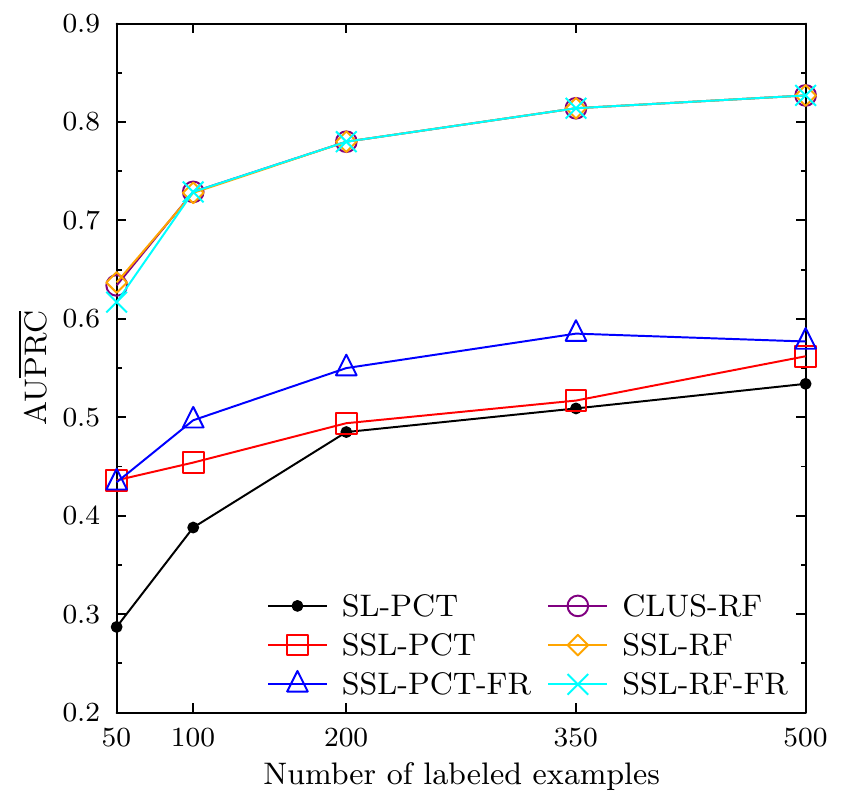}
	}
	\newline
	\subfloat[SIGMEA real\label{sigmea_pruned}]{%
		\includegraphics[height=0.3\textwidth]{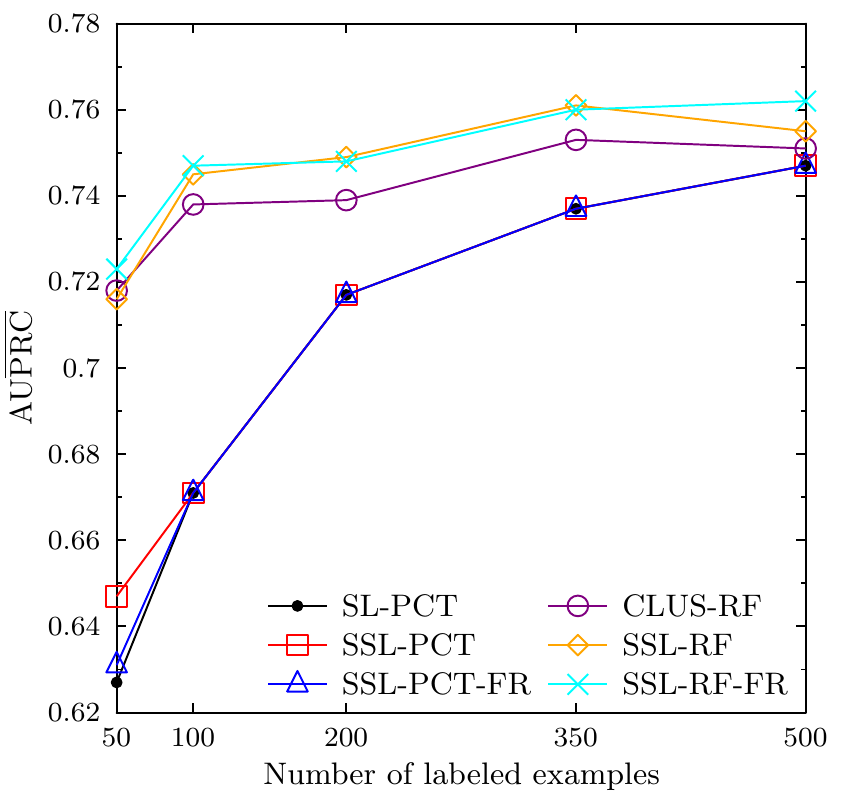}
	}
	\hfill
	\subfloat[Slovenian rivers\label{sloriv_pruned}]{%
		\includegraphics[height=0.3\textwidth]{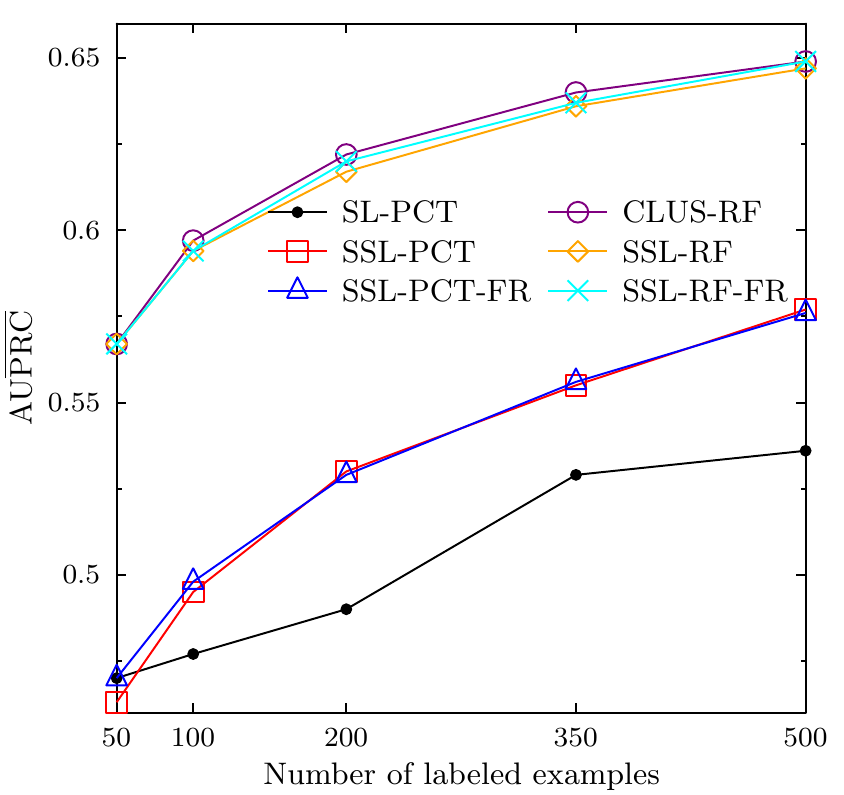}
	}
	\hfill
	\subfloat[Yeast\label{yeast_pruned}]{%
		\includegraphics[height=0.3\textwidth]{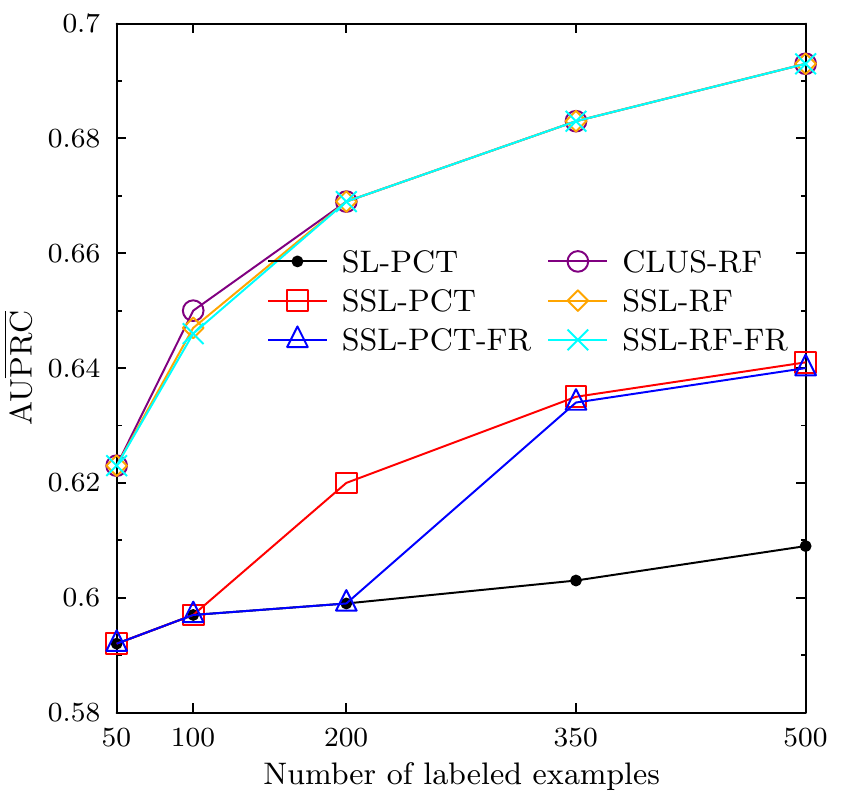}
	}
	\caption{\label{fig:results_sslpcts_mlc} Predictive performance of the supervised and semi-supervised methods on the multi-label classification datasets.}
\end{figure}

Figure~\ref{fig:results_sslpcts_mlc} presents the predictive performance ($\mathrm{AU}\overline{\mathrm{PRC}}$) of semi-supervised (\textsc{SSL-PCT}, \textsc{SSL-PTC-FR}, \textsc{SSL-RF} and \textsc{SSL-RF-FR}) and supervised methods (\textsc{SL-PCT} and \textsc{CLUS-RF}) on the 12 MLC datasets, with an increasing amount of labeled data.

We can clearly observe that semi-supervised PCTs are superior to \textsc{SL-PCT}s on most of the datasets. Namely, on 8 out of 12 datasets, either \textsc{SSL-PCT}s or \textsc{SSL-PCT-FR}s (or both) dominate the performance of \textsc{SL-PCT}s by a good margin. On the other four datasets, namely Corel5k, Emotions, Mediana and SIGMEA real, the performance of supervised and semi-supervised PCTs is mostly the same as, or similar to the performance of \textsc{SL-PCT}s.

Intuitively, the improvement of semi-supervised over supervised methods should diminish as the number of labeled examples increases, and eventually, semi-supervised and supervised methods are expected to converge to the same or similar performance. However, the ``convergence point'' changes from dataset to dataset. For instance, for Genbase, at 500 labeled examples we already see this convergence. For the other datasets, the improvement of semi-supervised over supervised methods decreases less quickly. 


The feature-weighted semi-supervised method (\textsc{SSL-PCT-FR}) and the non-feature-weighted one (\textsc{SSL-PCT}) have similar trends in predictive performance. However, on some datasets, there are notable differences. Namely, on Birds and Scene datasets, feature weighting is beneficial for the predictive performance of \textsc{SSL-PCT}s, and even necessary for improvement over \textsc{SL-PCT}s on the Birds dataset with $\geq$ 350 labeled examples. On the other hand, feature weighting clearly damages the predictive performance of the \textsc{SSL-PCT} method on the Bibtex dataset. Thus, feature weighting helps in most cases, but the empirical results cannot support its use by default when building \textsc{SSL-PCT}s for MLC. 

We next compare semi-supervised random forests (\textsc{SSL-RF}) with supervised random forests (\textsc{CLUS-RF}). From the results, we can observe that \textsc{CLUS-RF} improves over \textsc{CLUS-RF} on several datasets: Bibtex, Corel5k, Genbase, Medical, SIGMEA real, and marginally on Emotions and Enron datasets. However, as compared to single trees, the improvements of the semi-supervised approach over the supervised are observed on fewer datasets and are smaller in magnitude. In other words, the improvement of \textsc{SSL-PCT}s over \textsc{SL-PCT}s does not guarantee the improvement of \textsc{SSL-RF} over \textsc{CLUS-RF} (e.g., Mediana, and Yeast datasets), and vice versa, \textsc{SSL-RF} can improve over \textsc{CLUS-RF} even if \textsc{SSL-PCT}s does not improve over \textsc{SL-PCT}s (e.g., \textsc{SSL-RF-FR} on Emotions dataset for 200 and 350 labeled examples). As observed for the single trees, there is no clear advantage to using feature weighting when semi-supervised random forests are built, even though it is somewhat helpful on the Emotions and Enron datasets.

Feature-weighted PCTs and RFs could possibly be improved by considering a semi-supervised feature selection \cite{petkovic2022feature}, instead of the supervised method based on random forests, as used here.   

\subsubsection{Hierarchical multi-label classification}

\begin{figure}[!htbp]
	\subfloat[Danish farms\label{ecogen_pruned}]{%
		\includegraphics[height=0.3\textwidth]{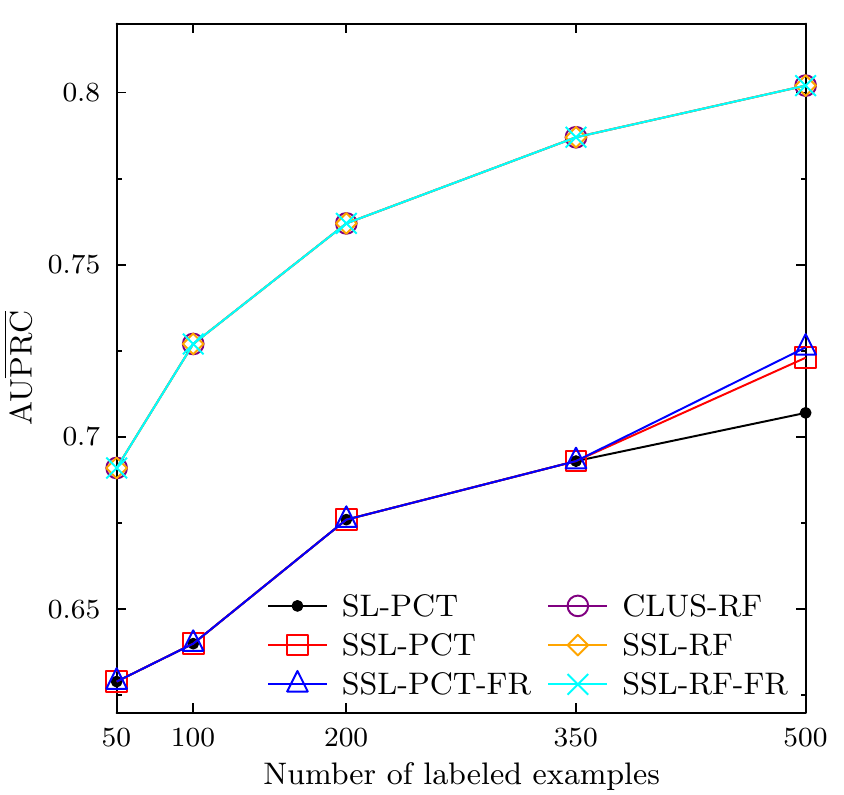}
	}
	\hfill
	\subfloat[Slovenian rivers\label{slo_riv_pruned}]{%
		\includegraphics[height=0.3\textwidth]{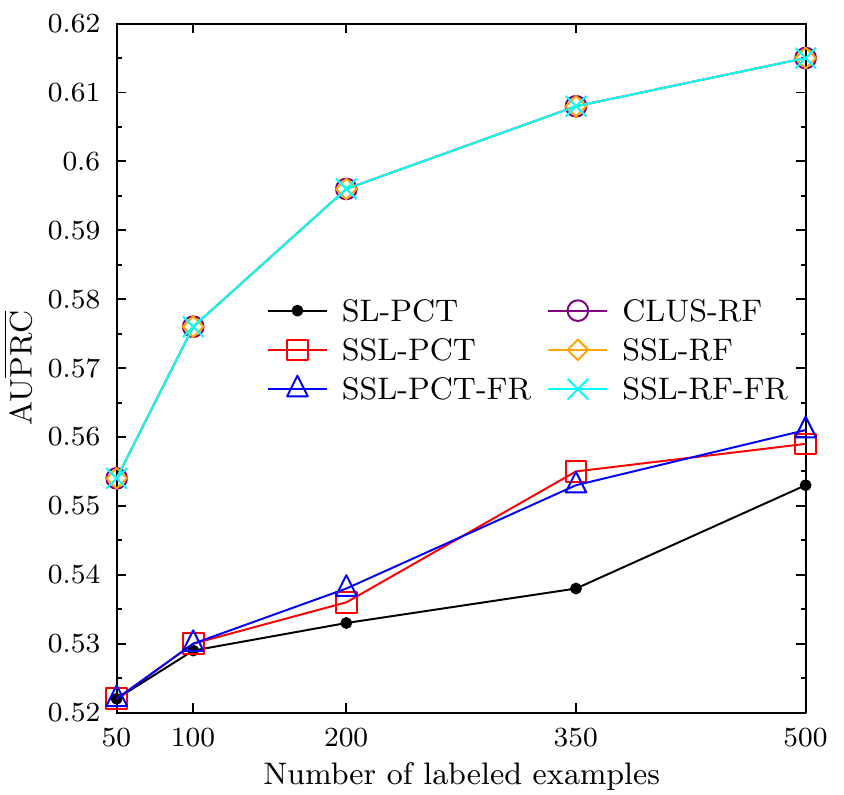}
	}
	\hfill
	\subfloat[Enron\label{enron_hmc_pruned}]{%
		\includegraphics[height=0.3\textwidth]{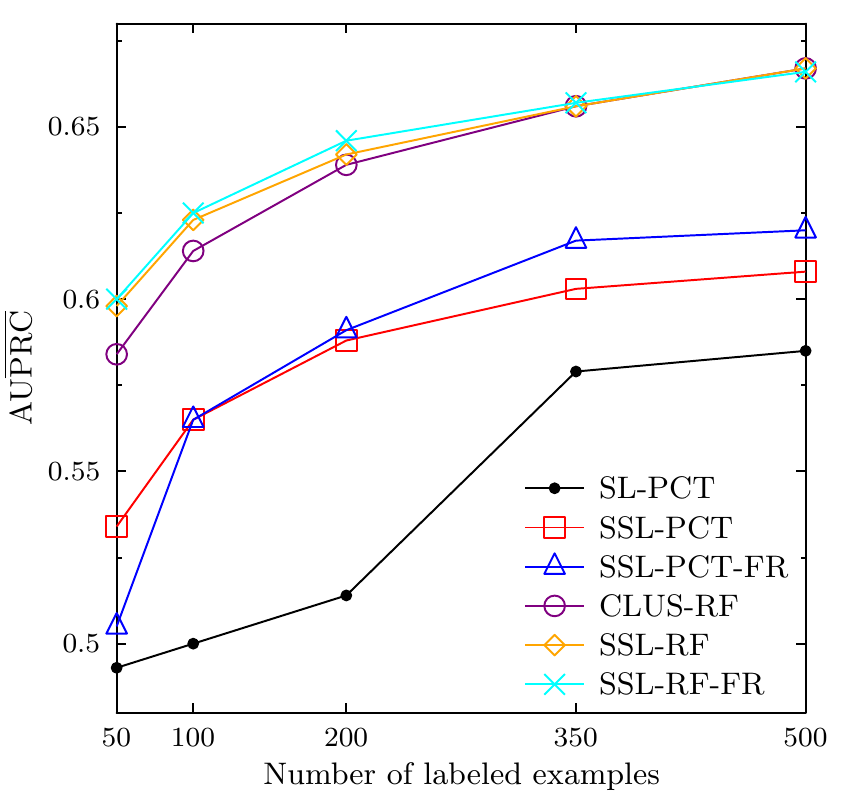}
	}
	\newline
	\subfloat[ImCLEF07A\label{ImCLEF07A_pruned}]{%
		\includegraphics[height=0.3\textwidth]{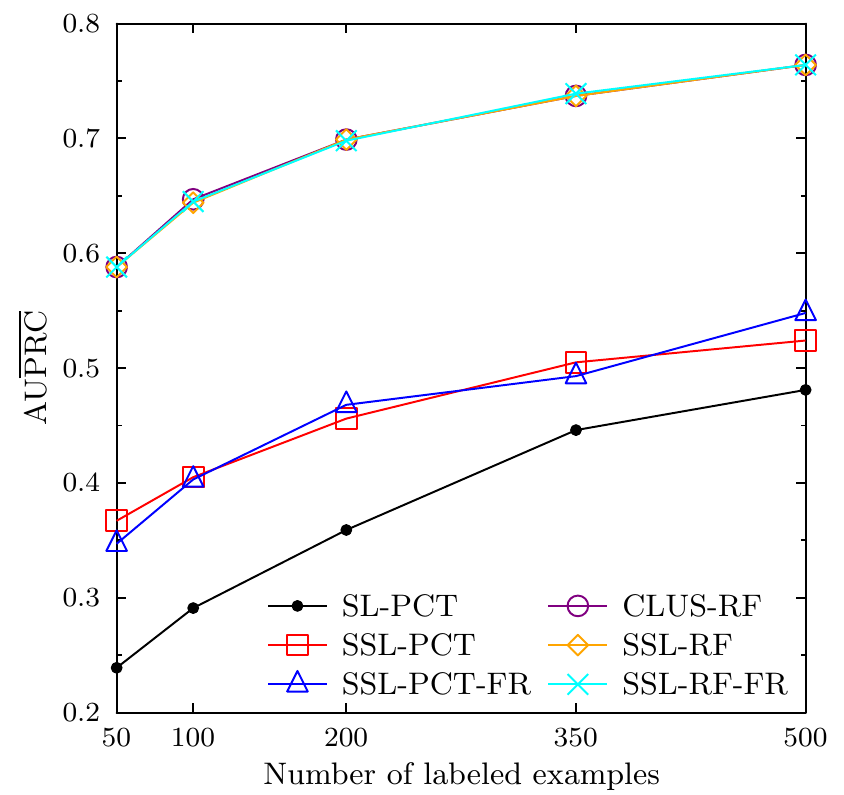}
	}
	\hfill
	\subfloat[ImCLEF07D\label{ImCLEF07D_pruned}]{%
		\includegraphics[height=0.3\textwidth]{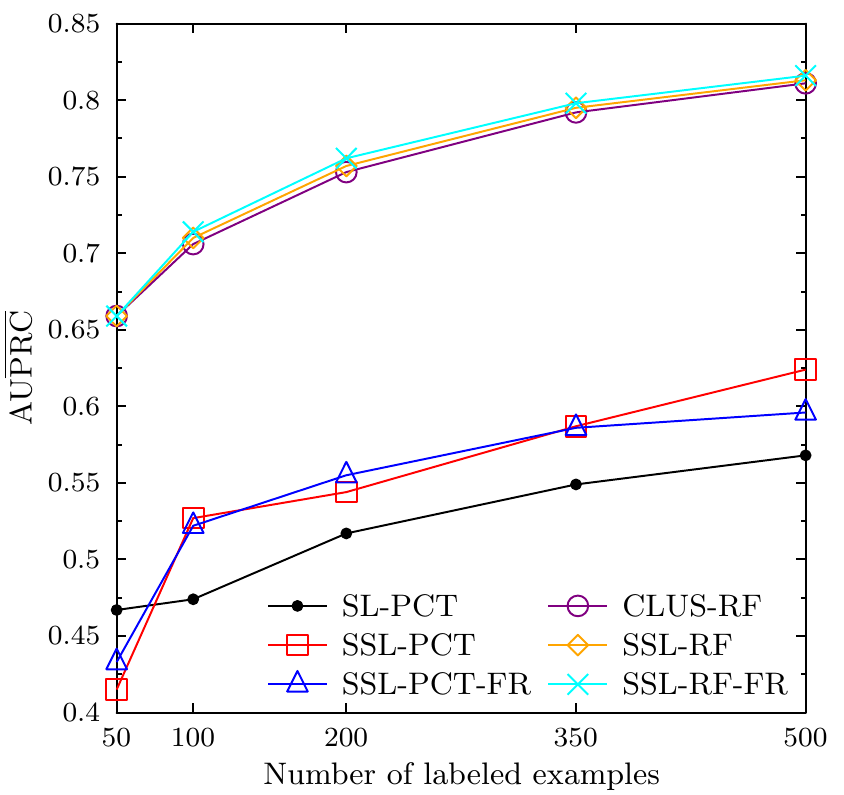}
	}
	\hfill
	\subfloat[Diatoms\label{Diatoms_pruned}]{%
		\includegraphics[height=0.3\textwidth]{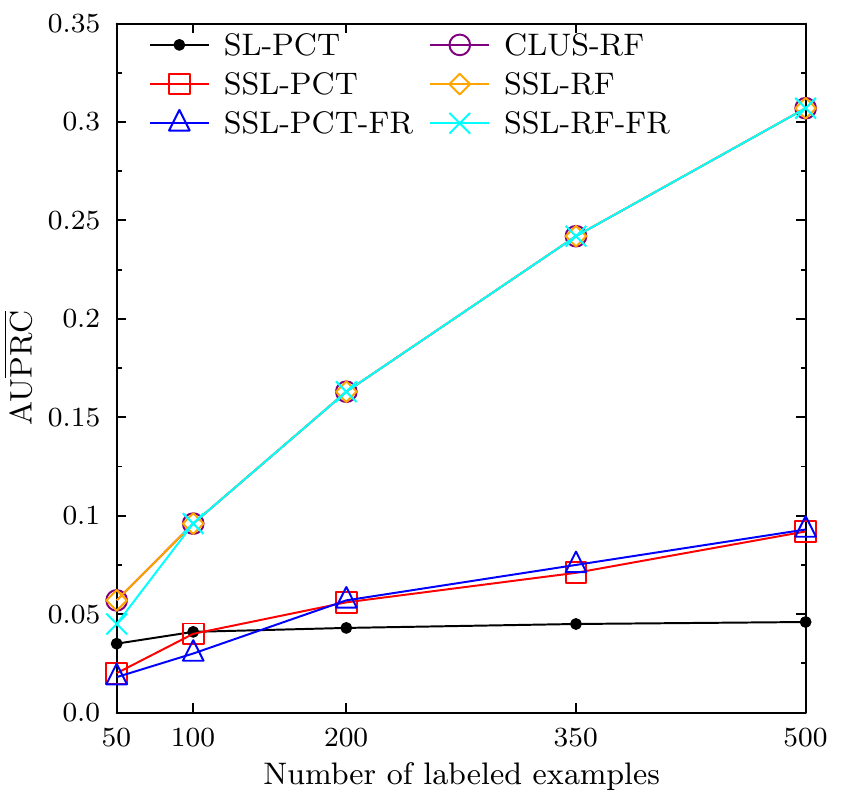}
	}
	\newline
	\subfloat[Cellcycle-GO\label{cellcycle_pruned}]{%
		\includegraphics[height=0.3\textwidth]{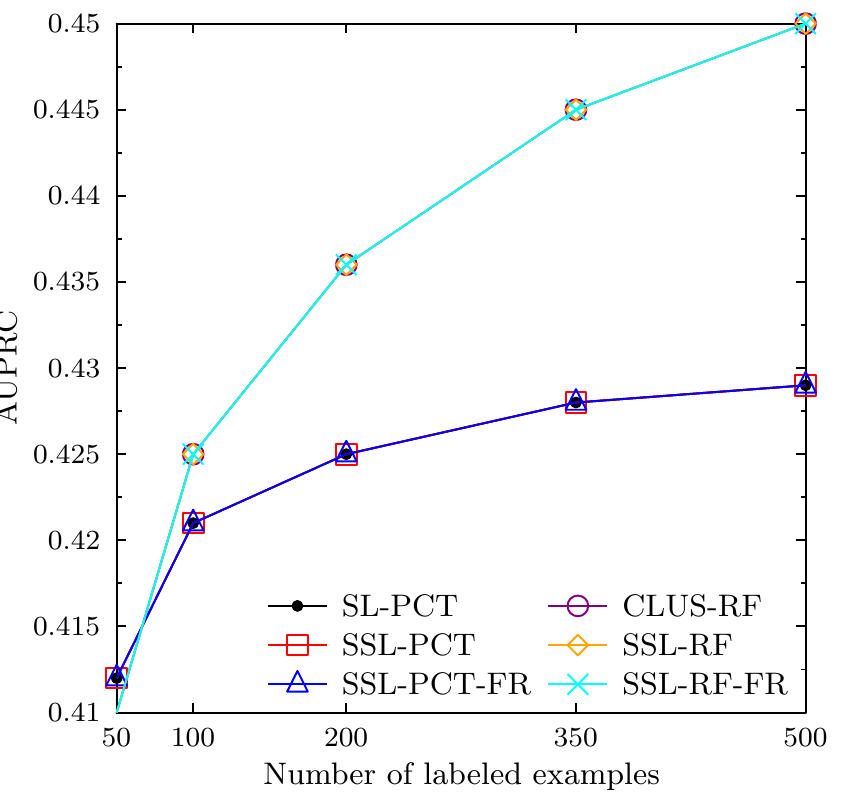}
	}
	\hfill
	\subfloat[Church-GO\label{church_pruned}]{%
		\includegraphics[height=0.3\textwidth]{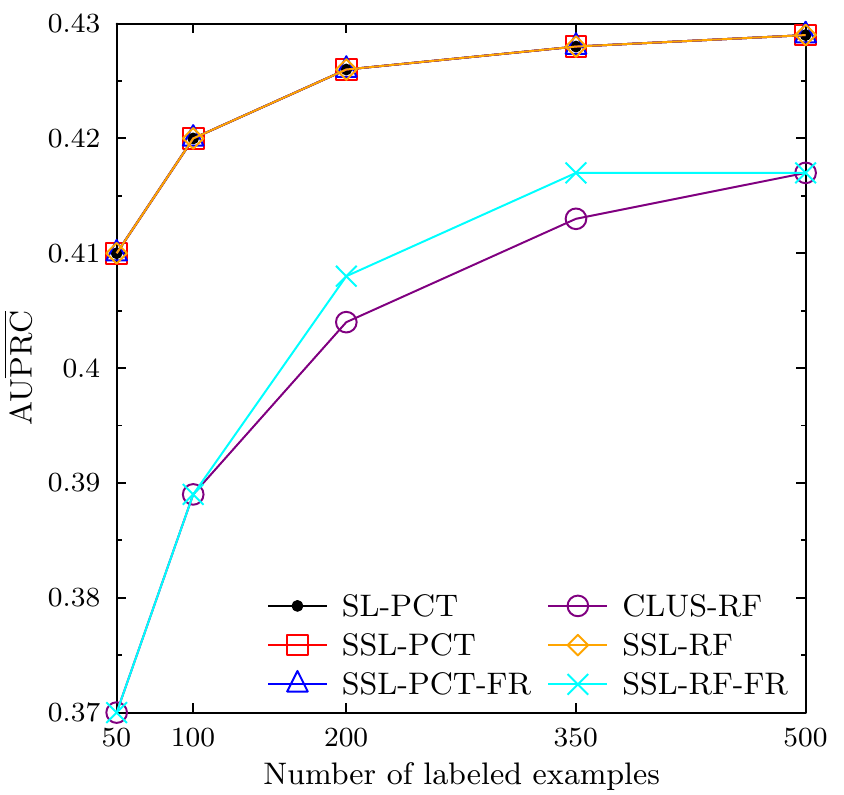}
	}
	\hfill
	\subfloat[Derisi-GO\label{derisi_pruned}]{%
		\includegraphics[height=0.3\textwidth]{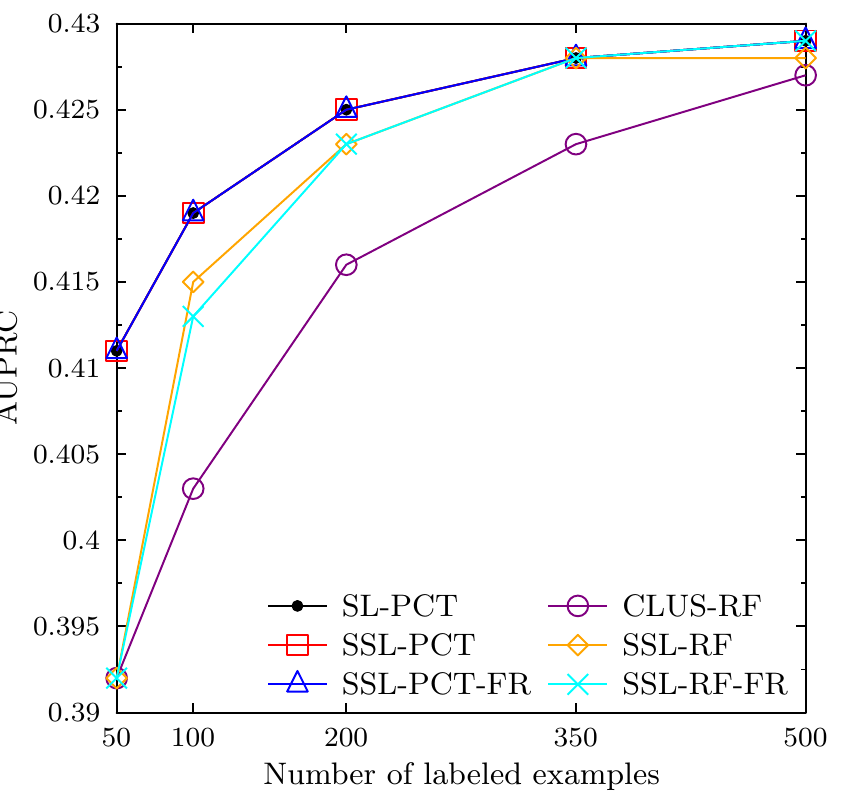}
	}
	\newline
	\subfloat[Eisen-GO\label{eisen_pruned}]{%
		\includegraphics[height=0.3\textwidth]{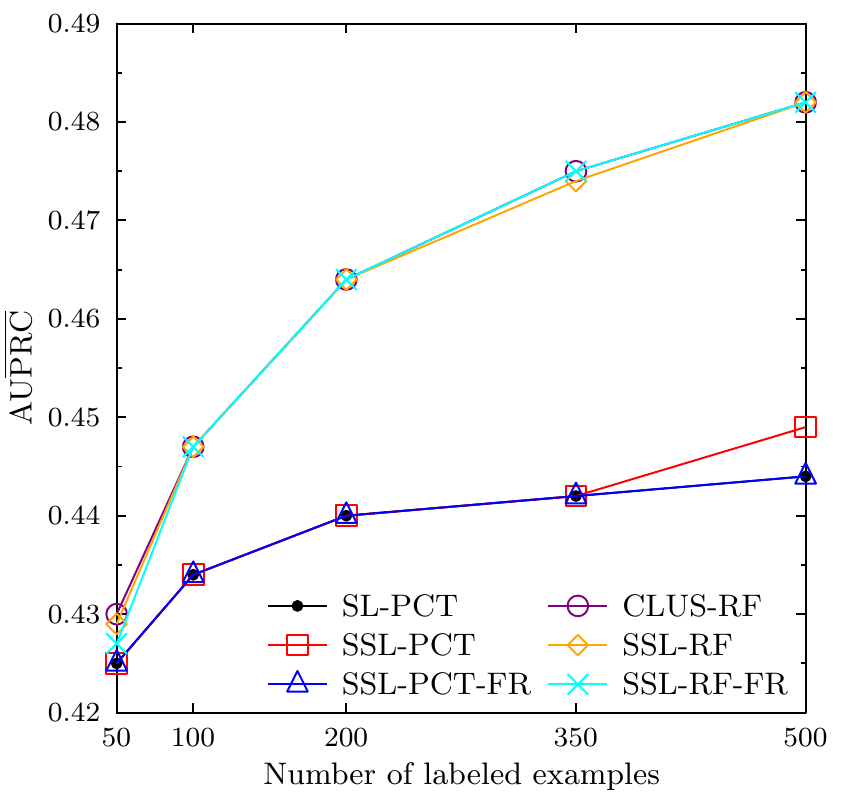}
	}
	\hfill
	\subfloat[Expr-GO\label{expre_pruned}]{%
		\includegraphics[height=0.3\textwidth]{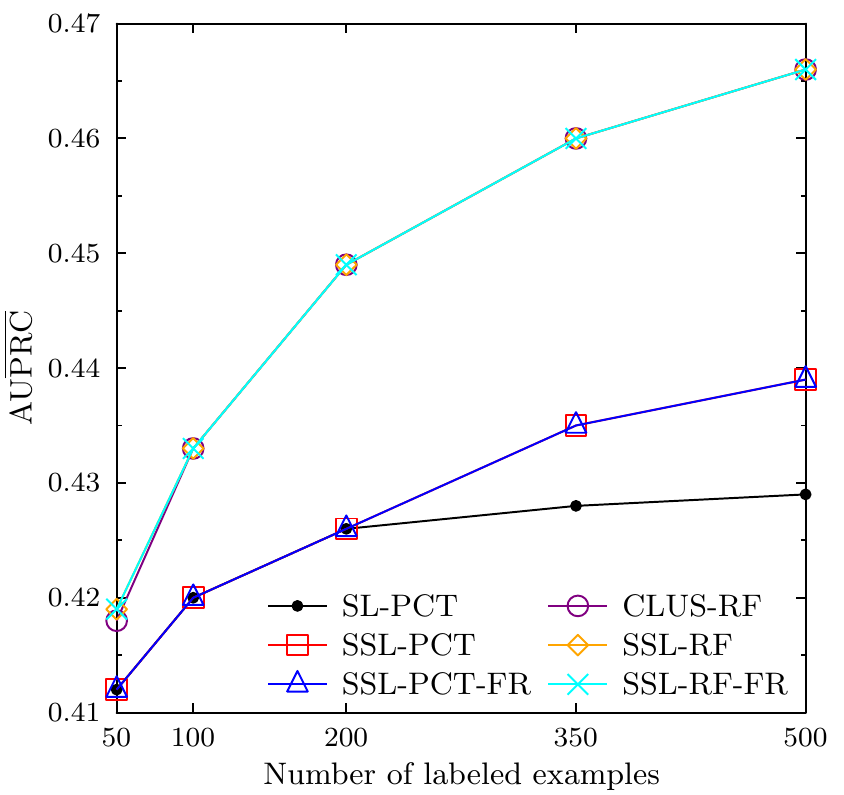}
	}
	\hfill
	\subfloat[Pheno-GO\label{pheno_pruned}]{%
		\includegraphics[height=0.3\textwidth]{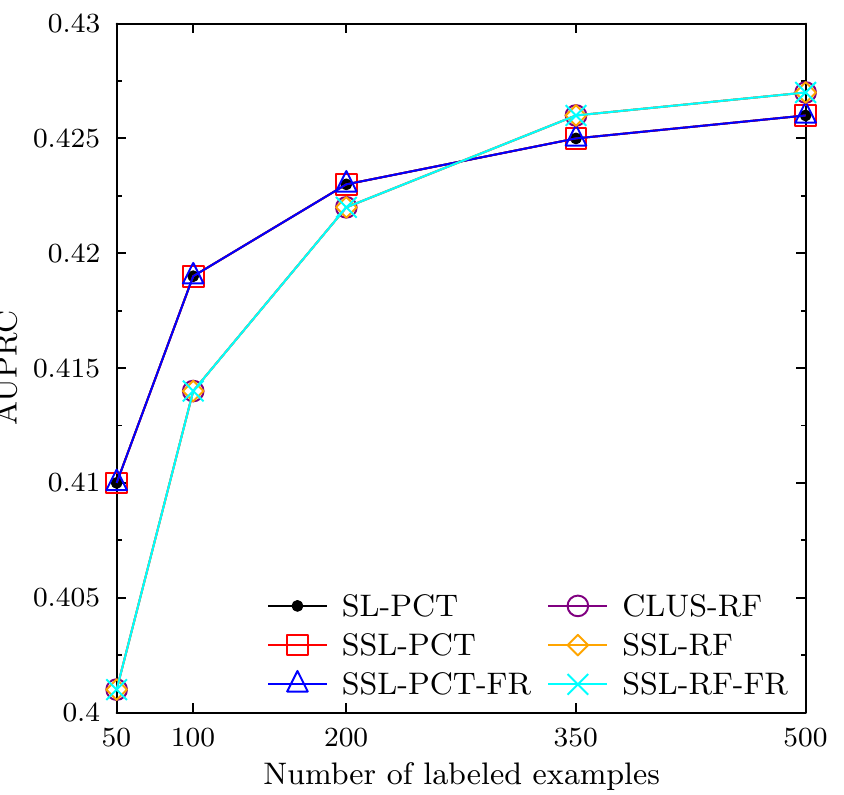}
	}
	\newline
	\caption{\label{fig:results_sslpcts_hmc} Predictive performance of the supervised and semi-supervised methods on the hierarchical multi-label classification datasets.}
\end{figure}

Figure~\ref{fig:results_sslpcts_hmc} presents the learning curves in terms of the predictive performance ($\mathrm{AU}\overline{\mathrm{PRC}}$) of semi-supervised (\textsc{SSL-PCT}, \textsc{SSL-PTC-FR}, \textsc{SSL-RF} and \textsc{SSL-RF-FR}) and supervised methods (\textsc{SL-PCT} and \textsc{CLUS-RF}) on the 12 hierarchical multi-label classification datasets.

We observe different behaviour on the 6 functional genomics datasets and the 6 datasets from the other domains. Namely, on all 6 datasets of the latter group, semi-supervised PCTs improve over \textsc{SL-PCT}s -- albeit not necessarily always for all available amounts of labeled data. This is the case on the Enron and ImCLEF07A datasets, where both \textsc{SSL-PCT}s and \textsc{SSL-PCT-FR} dominate the performance of \textsc{SL-PCT}s, while it seems that on other datasets at least 100 (Slovenian rivers and ImCLEF07D) or 500 (Danish farms) labeled examples are needed to improve over \textsc{SL-PCT}s. 

On the other hand, semi-supervised PCTs are not so successful on functional genomics datasets. Analysis of the tree sizes (see Section~\ref{sec:model_sizes}) reveals an explanation for such results. Namely, on all of the 6 functional genomics datasets, and for almost all different amounts of labeled data, both supervised and semi-supervised trees are composed of only one node. Note that, these datasets have extremely large label hierarchies which are very sparsely populated. It seems that for such datasets the amount of labeled data we considered (i.e., up to 500 labeled examples) is not sufficient to build trees -- neither supervised nor semi-supervised. In fact, semi-supervised trees have more than one node on Expr-GO ($\geq$350 of labeled examples) and Eisen (for 500 labeled examples) datasets, and those are exactly the occasions where they improve over supervised trees. We thus hypothesize that for larger amounts of labeled data, SSL-PCTs could outperform supervised PCT also on functional genomics datasets.

In the HMLC task the feature-weighted semi-supervised method (\textsc{SSL-PCT-FR}) and non-feature-weighted one (\textsc{SSL-PCT}) mostly have a very similar performance. Again, as for the MLC task, there is no clear benefit of feature weighting.

Finally, the semi-supervised random forests (\textsc{SSL-RF} and \textsc{SSL-RF-FR}) outperform supervised random forests (\textsc{CLUS-RF}) on some datasets, namely, in the initial part of the learning curve for the Enron dataset, and for the Church-GO and Derisi-GO datasets, meaning that unlabeled data improve the predictive performance of random forests of PCTs for HMLC. On the remaining datasets it seems that unlabeled data are not beneficial for the performance of random forests of PCTs for HMLC.

\subsection{Statistical analysis of predictive performance}

The results of the statistical analysis (Table~\ref{table:wilcoxon_sslpcts}) show that \textsc{SSL-PCT}s and \textsc{SSL-PCT-FR} are statistically significantly better than the \textsc{SL-PCT}s for most of the different amounts of labeled data, considered for both structured output prediction tasks. More specifically, for the HMLC task, usually, at least 200 labeled examples are needed to achieve statistical significance. In the MLC task, on the other hand, \textsc{SSL-PCT} achieves statistically significantly better results than \textsc{SL-PCT} up to 200 labeled examples. In this task, the feature-weighted SSL-PCTs are more successful: statistically, they significantly outperform \textsc{SL-PCT} across all different amounts of labeled examples.

Considering the feature-weighted and non-feature-weighted semi-supervised methods (both single trees and ensembles) there is no statistically significant difference between them in most cases, except at the HMLC task for 200 labeled examples where, statistically, \textsc{SSL-PCT-FR}  significantly outperforms \textsc{SSL-PCT}.

As discussed previously, semi-supervised random forests improve over supervised ones in fewer cases as compared to single trees. A statistically significant improvement over \textsc{CLUS-RF} is observed only for the MLC task with 200 labeled examples and the HMLC task with 350 labeled examples. However, in none of the cases, did the proposed semi-supervised methods perform statistically significantly worse than their supervised counterparts. 

\begin{table}[!tb]
	\centering
	\begin{footnotesize}
		\setlength\tabcolsep{3pt}
		\caption{{\textit P}-values of the Wilcoxon signed-rank test.}
		\label{table:wilcoxon_sslpcts}
		\begin{tabular}{lll | rrrrr}
			\hline
			\multicolumn{3}{c|}{Methods} & \multicolumn{5}{c}{Number of labeled examples} \\
			& & & \multicolumn{1}{c}{50} & \multicolumn{1}{c}{100} & \multicolumn{1}{c}{200} & \multicolumn{1}{c}{350} & \multicolumn{1}{c}{500} \\
            \hline
			\multicolumn{8}{c}{\textsc{Multi-label classification}} \\
			\hline
			SL-PCT & vs. & SSL-PCT & \textbf{0.012 $(+)$} & \textbf{0.008 $(+)$} & \textbf{0.008 $(+)$} & 0.117 $(+)$ & 0.071 $(+)$ \\
			SL-PCT & vs. & SSL-PCT-FR & \textbf{0.008 $(+)$} & \textbf{0.008 $(+)$} & \textbf{0.023 $(+)$} & \textbf{0.012 $(+)$} & \textbf{0.008 $(+)$} \\
			SSL-PCT & vs. & SSL-PCT-FR & 0.969 $(+)$ & 0.666 $(+)$ & 0.556 $(+)$ & 0.078 $(+)$ & 0.182 $(+)$ \\
			CLUS-RF & vs. & SSL-RF & 0.209 $(+)$ & 0.126 $(+)$ & 0.078 $(+)$ & 0.092 $(+)$ & 0.182 $(+)$ \\
			CLUS-RF & vs. & SSL-RF-FR & 0.17 $(+)$ & 0.117 $(+)$ & \textbf{0.013 $(+)$} & 0.638 $(+)$ & 0.695 $(-)$ \\
			SSL-RF & vs. & SSL-RF-FR & 0.937 $(+)$ & 0.209 $(+)$ & 0.754 $(-)$ & 0.327 $(-)$ & 0.388 $(-)$ \\
			\hline
			\multicolumn{8}{c}{\textsc{Hierarchical multi-label classification}} \\
			\hline
			SL-PCT & vs. & SSL-PCT & 0.937 $(+)$ & 0.147 $(+)$ & \textbf{0.034 $(+)$} & \textbf{0.025 $(+)$} & \textbf{0.008 $(+)$} \\
			SL-PCT & vs. & SSL-PCT-FR & 1 $(-)$ & 0.158 $(+)$ & \textbf{0.034 $(+)$} & \textbf{0.025 $(+)$} & \textbf{0.01 $(+)$} \\
			SSL-PCT & vs. & SSL-PCT-FR & 0.388 $(-)$ & 0.136 $(-)$ & \textbf{0.034 $(+)$} & 0.695 $(-)$ & 0.347 $(+)$ \\
			CLUS-RF & vs. & SSL-RF & 0.367 $(+)$ & 0.136 $(+)$ & 0.099 $(+)$ & 0.347 $(+)$ & 0.136 $(+)$ \\
			CLUS-RF & vs. & SSL-RF-FR & 1 $(-)$ & 0.347 $(+)$ & 0.136 $(+)$ & \textbf{0.034 $(+)$} & 0.48 $(+)$ \\
			SSL-RF & vs. & SSL-RF-FR & 0.367 $(-)$ & 0.666 $(+)$ & 1 $(-)$ & 0.239 $(+)$ & 0.969 $(-)$ \\				
			\hline
		\end{tabular}
        \\[5pt]
        \caption*{The statistical test is applied to the predictive performances ($\mathrm{AU}\overline{\mathrm{PRC}}$) of the supervised and semi-supervised single trees (\textsc{SL-PCT}, \textsc{SSL-PCT} and \textsc{SSL-PCT-FR}) on the datasets considered in this study: 12 for multi-label classification and 12 for hierarchical multi-label classification. In bold we report significant $p$-values ($<$ 0.05). In a comparison of the two algorithms, i.e., Algorithm1 vs. Algorithm2, the '$-$' sign indicates that the sum of ranks where the first algorithm outperformed the second is higher than the sum of ranks where the second algorithm outperformed the first. The '$+$' sign indicates the opposite.}
	\end{footnotesize}
\end{table}

\subsection{Influence of the amount of supervision}\label{sec:wparam_sslpct_mtr}

\begin{figure}[!htb]
	\centering
	\setlength\tabcolsep{3pt}
	\begin{tabular}{c}
		\includegraphics[width=0.9\textwidth]{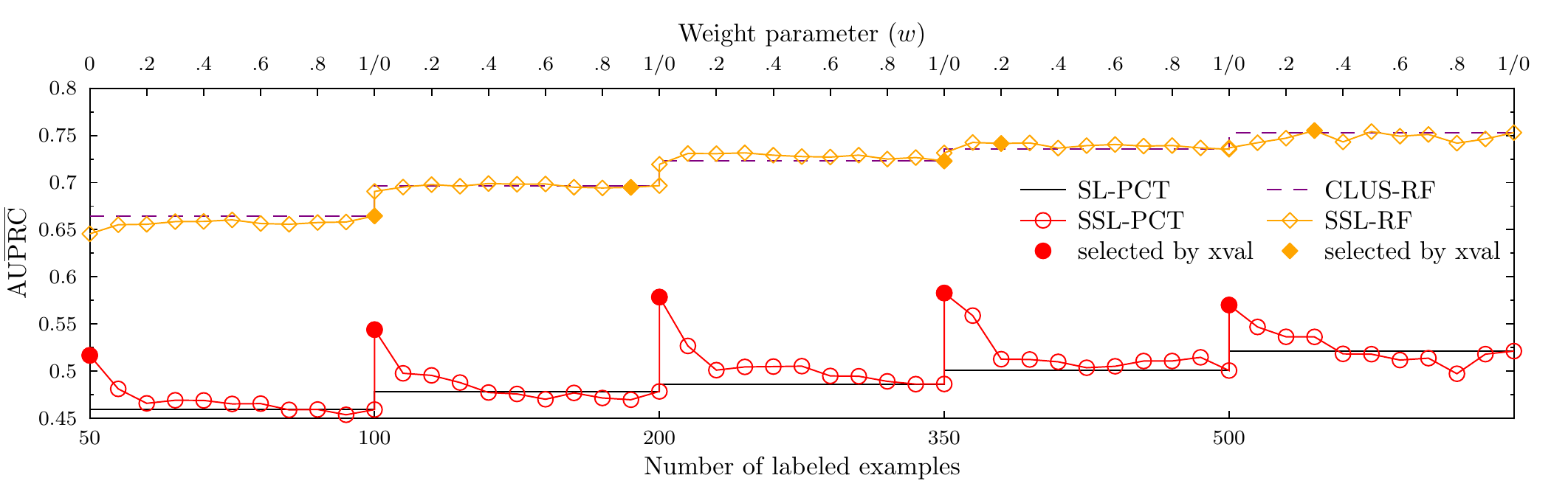} \\
		(a) Emotions (MLC) \\
        \includegraphics[width=0.9\textwidth]{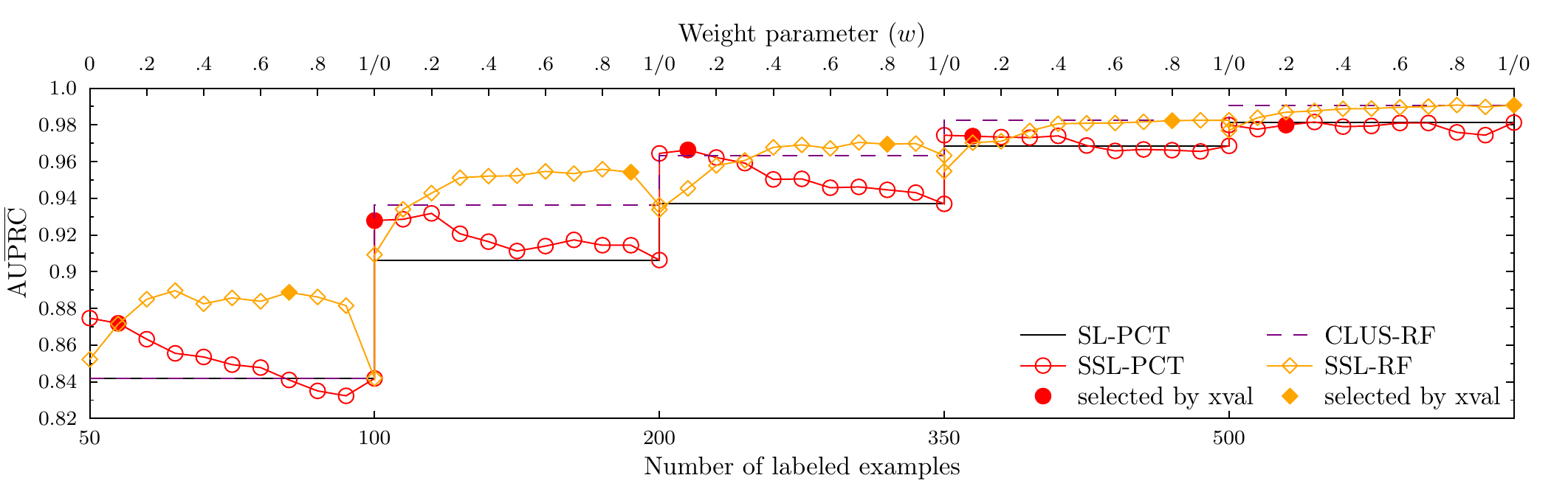} \\
		(b) Genbase (MLC) \\
		\includegraphics[width=0.9\textwidth]{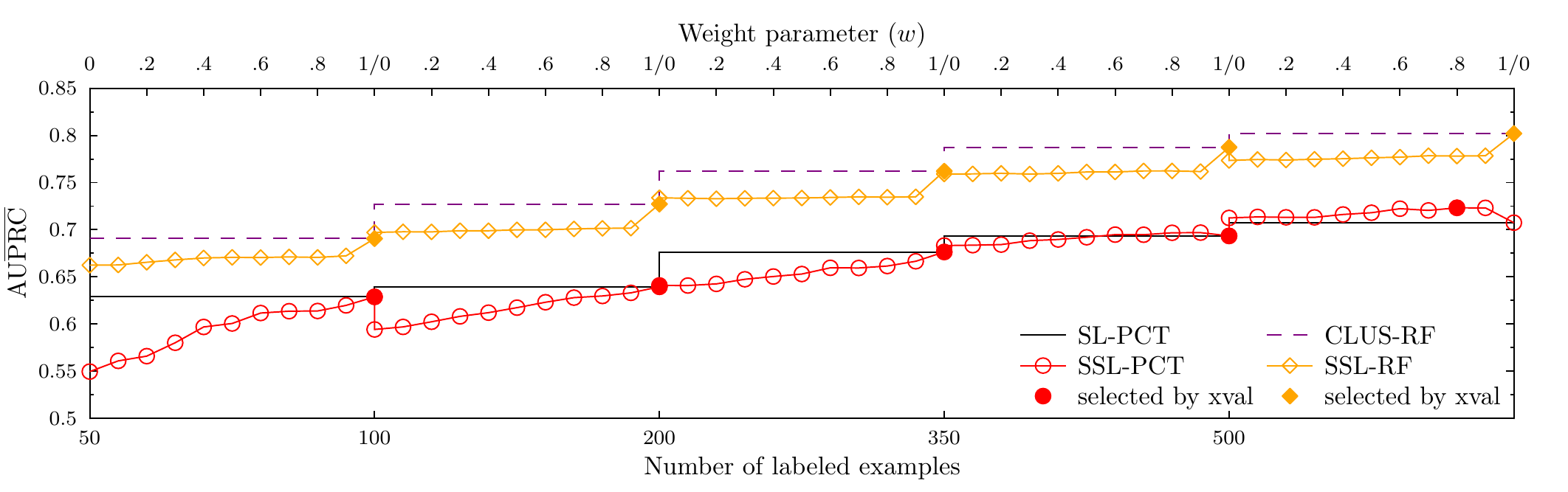} \\
		(c) Danish farms (HMLC) \\
        \includegraphics[width=0.9\textwidth]{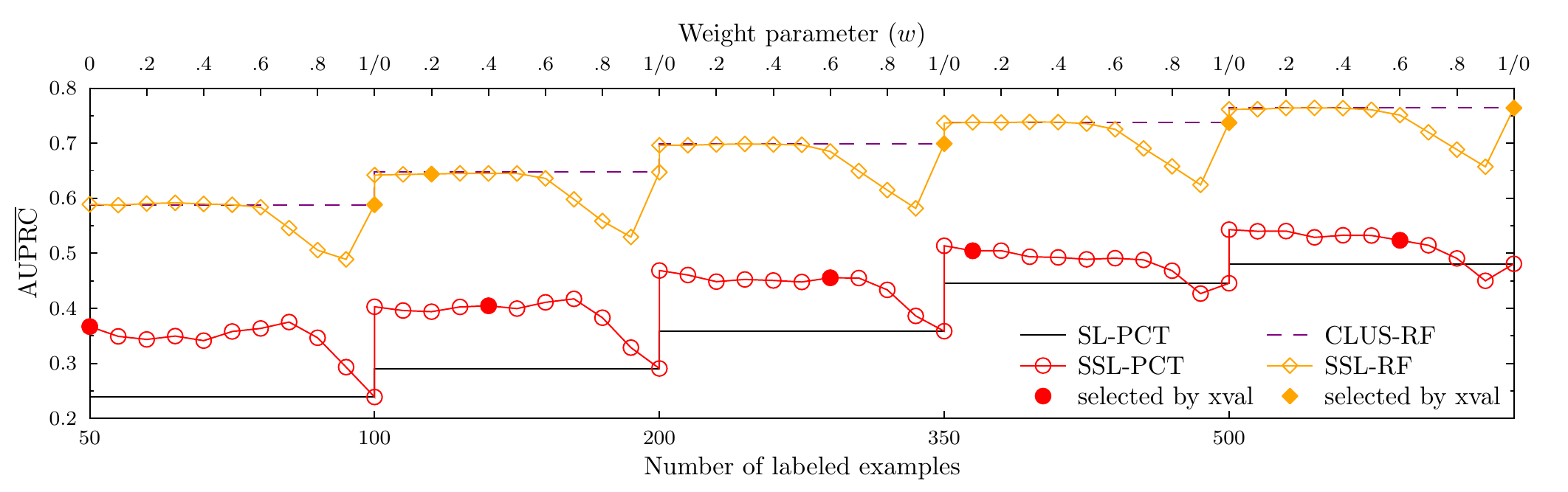} \\
		(d) ImCLEF07A (HMLC) \\
	\end{tabular}
	\caption{Influence of parameter $w$ on \textsc{SSL-PCT} (red line) and \textsc{SSL-RF} (orange line) methods. The results refer to 4 datasets with different types of structured outputs: Emotions (MLC), Genbase (MLC), Danish farms (HMLC), and ImCLEF07A (HMLC). The $w$ values selected by the internal cross-validation algorithm and used in the experiments are marked with colored dots.}
	\label{fig:results_weights}
\end{figure}

As previously mentioned, the amount of supervision in the SSL-PCTs is controlled by the $w$ parameter, where $w=0$ results in unsupervised PCTs, $0<w<1$ in semi-supervised PCTs, and $w=1$ in supervised PCTs. This ability to tune the degree of supervision in \textsc{SSL-PCT}s for the predictive problem at hand is of great practical importance. Namely, semi-supervised methods can, in general, degrade the performance of their supervised counterparts \cite{nigam2000text,cozman_unlabeled_2002,zhou_semisupervised_2007,guo_extensive_2010}. In this respect, some studies noted that the success of semi-supervised methods is domain-dependent \cite{chawla_learning_2005}. How to choose a suitable SSL method for the dataset at hand is an unresolved issue; therefore, even if the primary task of SSL methods is to improve the performance over supervised methods, making semi-supervised methods \textit{safe} is also a high priority. In other words, making sure that SSL methods do not degrade the performance of the corresponding supervised methods.

In SSL-PCTs, such a safety mechanism is provided by the $w$ parameter. Theoretically, given the optimal value of $w$, \textsc{SSL-PCT}s and \textsc{SSL-RF} would always perform at least as well as their supervised counterparts both for MLC and HMLC. The reason is that \textsc{SL-PCT}s and \textsc{CLUS-RF} are special cases of \textsc{SSL-PCT} and \textsc{SSL-RF} when $w=1$. In practice, however, the $w$ parameter is chosen via internal cross-validation on labeled examples in the training set. Thus, it is possible to select $w$ sub-optimal for the test set considered.

Our empirical evaluation showed that, \textsc{SSL-PCT} and \textsc{SSL-RF} rarely degrade the performance of \textsc{SL-PCT} and \textsc{CLUS-RF} (Figures~\ref{fig:results_sslpcts_mlc} and \ref{fig:results_sslpcts_hmc}). Across all the experiments we performed, \textsc{SSL-PCT}s outperformed their supervised counterpart (\textsc{SL-PCT}) in 52\% of the experiments, performed worse in 9\% of the experiments, and performed equally in 39\% of the experiments. 
Moreover, the occasional degradation of the predictive performance was small compared to the improvement of \textsc{SSL-PCT} over \textsc{SL-PCT}. For example, the average relative improvement of \textsc{SSL-PCT}s over \textsc{SL-PCT}s (across all the experiments) was $40\%$, while the average degradation was $7\%$.

Figure~\ref{fig:results_weights} clearly shows the role of parameter $w$ on the predictive performance for 4 datasets with different types of structured output. 
The Emotions dataset (Fig.~\ref{fig:results_weights}a) requires no supervision because $w=0$ provides better predictive performance of the \textsc{SSL-PCT} method, whereas the Genbase dataset (Fig.~\ref{fig:results_weights}b) requires small amount of supervision (i.e., $w$ close to 0) for \textsc{SSL-PCT} and high amount of supervision (i.e., $w$ close to 1) for \textsc{SSL-RF}. For the HMLC dataset, Danish Farms (Fig.~\ref{fig:results_weights}c), more supervision (i.e., higher $w$) provides better predictive performance of \textsc{SSL-PCT}. However, for up to 500 labeled examples, the \textsc{SSL-PCT} method is unable to improve its supervised counterpart; therefore, $w=1$ is selected to prevent performance degradation. For the other HMLC dataset, ImCLEF07A (Fig.~\ref{fig:results_weights}d), on the other hand, the performance drops for high levels of supervision (i.e., $w>0.5$). 

In conclusion, our results show that the optimal value of $w$ depends on the dataset and on the different amounts of labeled data, as exemplified in Figure~\ref{fig:results_weights}. This confirms our initial intuition that a different amount of supervision is suitable for different datasets. Therefore, it is difficult to provide a general recommendation for the value of $w$, and it is advisable to optimize this parameter by internal cross-validation for each dataset, as it is done in our study.

\subsection{Interpretability of the models}\label{sec:model_sizes}

\begin{figure}[!b]
	\subfloat[Supervised PCT\label{tree_sl}]{%
		\includegraphics[height=0.29\textwidth]{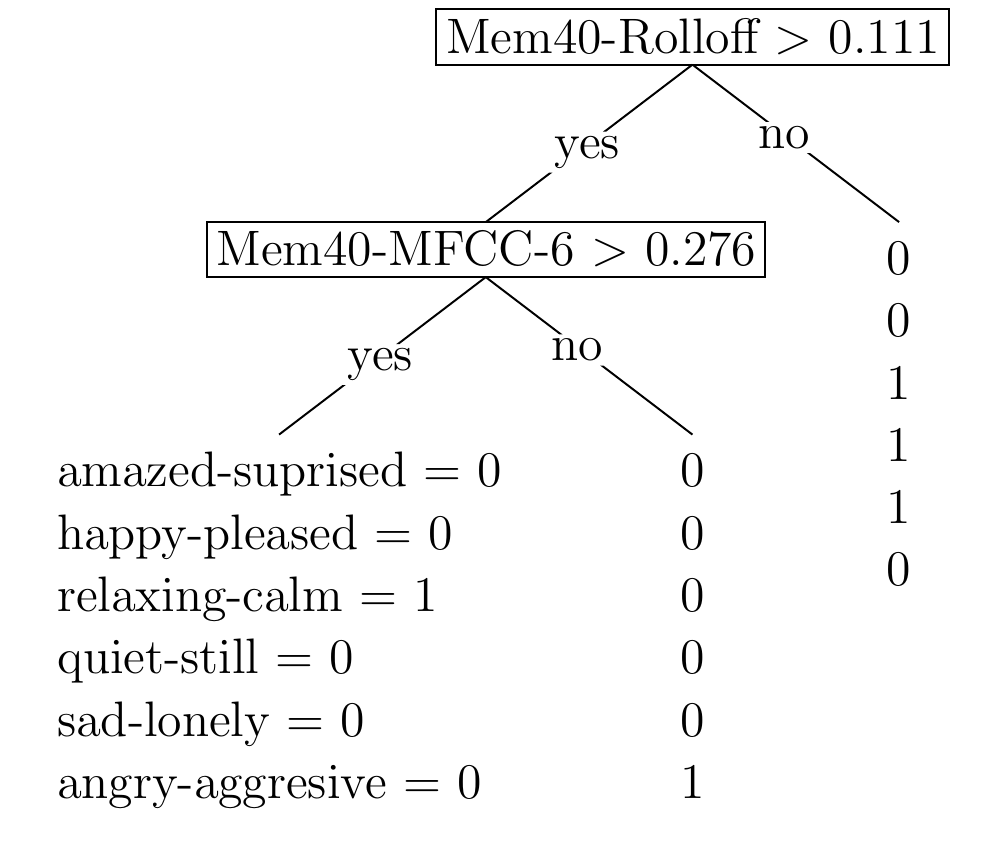}
	}
	\hfill
	\subfloat[Semi-supervised PCT\label{tree_ssl}]{%
		\includegraphics[height=0.29\textwidth]{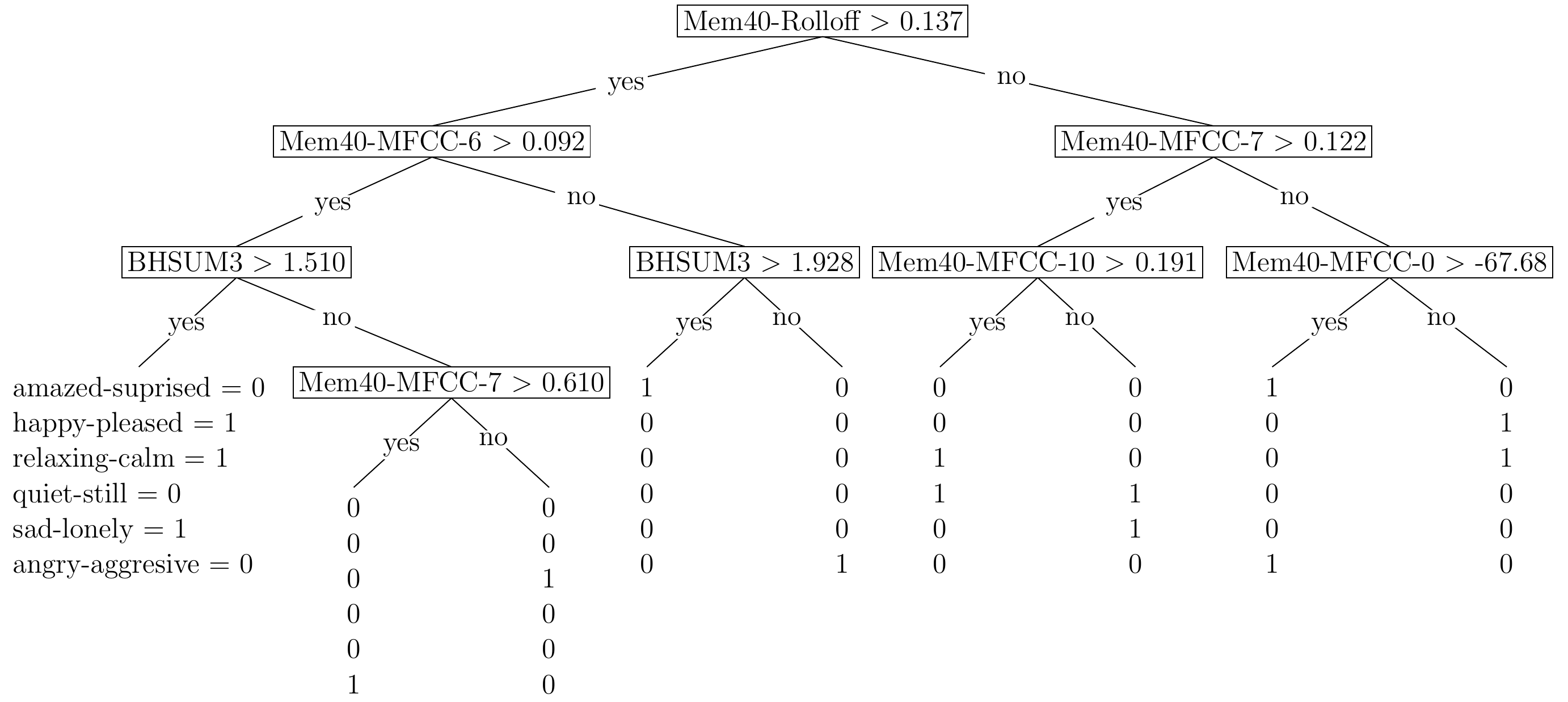}
	}
	\caption{\label{fig:trees} Supervised and semi-supervised predictive clustering trees obtained for the Emotions dataset with 100 labeled examples.}
\end{figure}

Interpretability of the predictive models is often a desirable property of machine learning algorithms. Since the models produced by the SSL-PCTs are in the form of a decision tree, they are readily interpretable. To the best of our knowledge, in the literature, no other semi-supervised method for MLC and HMLC produces interpretable models.

The degree of interpretability of the tree-based models is typically expressed in terms of their size. A large tree can be more difficult to interpret, and vice versa, a small tree can be easier to interpret. The tree size is often a trade-off between accuracy and interpretability. Small trees are easy to interpret, but due to their simplicity may fail to capture interactions in the data and therefore provide a satisfactory accuracy. On the other hand, larger trees may mitigate such issues, but at the cost of lower interpretability. Note that increased size does not necessarily mean improved predictive power of tree models, due to possible overfitting. In general, it is not easy to identify (\textit{a priori}) the best size of a tree, in order to balance between overfitting and underfitting.

\begin{table}[!bt]
	\centering
	\begin{footnotesize}
		\setlength\tabcolsep{4pt}
		\caption{Model sizes expressed as the number of nodes in trees.}
		\label{table:sslpcts_sizes}
		\begin{tabular}{l | rr|rr|rr|rr|rr}
			\hline
			\multirow{4}{*}{Dataset} & \multicolumn{10}{c}{Number of labeled examples} \\
			& \multicolumn{2}{c}{50} & \multicolumn{2}{c}{100} & \multicolumn{2}{c}{200} & \multicolumn{2}{c}{350} & \multicolumn{2}{c}{500} \\ \cline{2-11}
			& SL & SSL & SL & SSL & SL & SSL & SL & SSL & SL & SSL \\
			& -PCT & -PCT & -PCT & -PCT & -PCT & -PCT & -PCT & -PCT & -PCT & -PCT \\ 
            \hline
			\multicolumn{11}{c}{\textsc{Multi-label classification}} \\
			\hline
			Bibtex & 1 & 21 & 1 & 21.4 & 1 & 19.4 & 1 & 19.4 & 1 & 19 \\
			Birds & 1 & 15.8 & 1 & 20.8 & 1.2 & 15.2 & 2.6 & 13 & 4 & 12.8 \\
			Corel5k & 1 & 33.6 & 1 & 1 & 1 & 1 & 1 & 1 & 1 & 1 \\
			Emotions & 3 & 18.8 & 5 & 19 & 7.4 & 7.4 & 11.8 & 19.2 & 14.8 & 14.8 \\
			Enron & 1 & 17.6 & 1 & 19.2 & 1 & 21.4 & 1 & 21.2 & 1.2 & 20.2 \\
			Genbase & 2.6 & 22.8 & 21.4 & 23 & 31.2 & 26.6 & 37.2 & 36.8 & 43 & 41.4 \\
			Mediana & 1.4 & 32.8 & 4 & 4 & 7.8 & 7.8 & 10.2 & 10.2 & 12.4 & 12.4 \\
			Medical & 1 & 15.4 & 1 & 41.8 & 1 & 43.4 & 3.2 & 63.6 & 13.6 & 63.2 \\
			Scene & 7.8 & 25.8 & 12.4 & 28.4 & 19.8 & 28.4 & 31.8 & 29 & 36.6 & 37.2 \\
			SIGMEA real & 3 & 35.2 & 3.2 & 3.2 & 4.6 & 4.6 & 6.6 & 6.6 & 8.4 & 8.4 \\
			Slovenian rivers & 1 & 39.2 & 1 & 64 & 1.4 & 67.6 & 3 & 70.8 & 3.8 & 55.8 \\
			Yeast & 1 & 1 & 1 & 1 & 1 & 25 & 1.2 & 25 & 2.2 & 25 \\
			\multicolumn{1}{r|}{\textbf{Average:}} & \textbf{2.1} & \textbf{23.3} & \textbf{4.4} & \textbf{20.6} & \textbf{6.5} & \textbf{22.3} & \textbf{9.2} & \textbf{26.3} & \textbf{11.8} & \textbf{25.9} \\
			\hline
			\multicolumn{11}{c}{\textsc{Hierarchical multi-label classification}} \\
			\hline
			Danish farms & 1 & 1 & 1.4 & 1.4 & 3.2 & 3.2 & 6 & 6 & 8.8 & 257.6 \\
			Slovenian rivers & 1 & 19.8 & 1 & 18.2 & 1 & 47.4 & 1.4 & 52.4 & 3 & 50.8 \\
			Enron & 1 & 11.6 & 1 & 13.8 & 1.6 & 15.8 & 5.6 & 16.6 & 6.6 & 17.2 \\
			ImCLEF07A & 1 & 42.8 & 2 & 86.6 & 6.6 & 133.4 & 12.8 & 150.4 & 19.4 & 177.4 \\
			ImCLEF07D & 1 & 47.4 & 1.8 & 215 & 4.4 & 295.2 & 7 & 159.2 & 15 & 172.2 \\
			Diatoms & 1 & 49.6 & 1 & 56.2 & 1 & 57.4 & 1 & 72.4 & 1 & 89.8 \\
			Cellcycle-GO & 1 & 1 & 1 & 1 & 1 & 1 & 1 & 1 & 1 & 1 \\
			Church-GO & 1 & 1 & 1 & 1 & 1 & 1 & 1 & 1 & 1 & 1 \\
			Derisi-GO & 1 & 1 & 1 & 1 & 1 & 1 & 1 & 1 & 1 & 1 \\
			Eisen-GO & 1 & 1 & 1 & 1 & 1 & 1 & 1 & 1 & 1 & 27.4 \\
			Expr-GO & 1 & 1 & 1 & 1 & 1 & 1 & 1 & 9.2 & 1 & 9.2 \\
			Pheno-GO & 1 & 1 & 1 & 1 & 1 & 1 & 1 & 1 & 1 & 1 \\
			\multicolumn{1}{r|}{\textbf{Average:}} & \textbf{1.0} & \textbf{14.9} & \textbf{1.2} & \textbf{33.1} & \textbf{2.0} & \textbf{46.5} & \textbf{3.3} & \textbf{39.3} & \textbf{5.0} & \textbf{67.1} \\			
			\hline			
		\end{tabular}
	\end{footnotesize}
\end{table}

In Table~\ref{table:sslpcts_sizes}, we compare tree sizes of supervised and semi-supervised PCTs. We observe that, on average, the semi-supervised trees are somewhat larger than the supervised trees. This is intuitive since semi-supervised algorithms use much more data to grow the trees, i.e., both labeled and unlabeled examples. If we focus on individual datasets, we can observe that the size of both the supervised and semi-supervised trees is mainly in the range of a few tens of nodes. This is still a reasonable size for manual inspection. However, there are a few exceptions. Semi-supervised trees are sometimes, with a few hundred nodes, much larger than the corresponding supervised trees. In particular, this can be observed in the following datasets: Mediana ($\geq$ 350 labeled), Danish farms (500 labeled), ImCLEF07A, and ImCLEF07D ($\geq$ 200 labeled). These cases, generally characterized by a large number of classes, can be infeasible for analysis.

To exemplify the interpretability and to highlight the possible differences between SL-PCTs and SSL-PCTs, we provide an example of supervised and semi-supervised predictive clustering trees obtained for the Emotions dataset with 100 labeled examples (Figure~\ref{fig:trees}) where the task is to predict an emotion evoked by music on the basis of features such as Mel Frequency Cepstral Coefficients (MFCCs) that describe timbre or Rolloff describing a frequency response below or above a certain limit. We can observe that unlabeled examples enabled the semi-supervised algorithm to build a larger and, in this case, more accurate tree than the supervised one (note the predictive performance in Figure~\ref{fig:results_sslpcts_mlc}). Next, we can observe that the most important features (i.e., the ones at the top of the tree) are the same in both trees, however, the splitting points are different implying that unlabeled examples can help semi-supervised trees to refine the splits. 

\begin{figure}[!tb]
	\centering
		\includegraphics[height=0.35\textwidth]{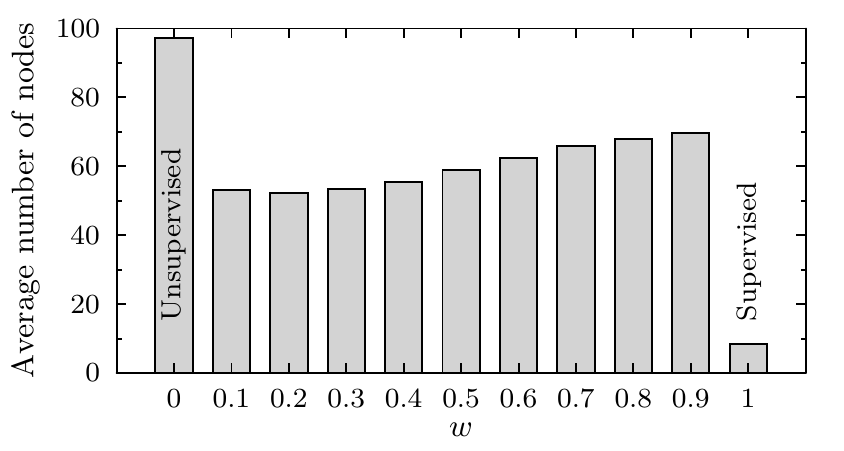}
	\caption{\label{fig:sslpcts_avgSize} Average tree size per value of parameter $w$ across all datasets and amounts of labeled data.}
\end{figure}

A closer analysis of the results is shown in Figure~\ref{fig:sslpcts_avgSize}, where it is possible to evaluate the influence of parameter $w$ on the tree size. The analysis reveals that unsupervised trees ($w=0$) are much bigger than semi-supervised ($0 < w < 1$) or supervised ($w=1$) trees. Unsupervised trees do not rely on the output space at all, therefore, it is understandable that, in the presence of a very large amount of unlabeled data, big trees are grown. We recall that the $W$ parameter is optimized for predictive performance, but by increasing the value of $w$ (i.e., increasing the degree of supervision) a trade-off between tree size and model performance can be achieved. 

\subsection{Training times}\label{sec:model_times}

In Table \ref{table:sslpcts_time} we present the training times of supervised and semi-supervised algorithms. For simplicity, we present times for experiments with 500 labeled examples, since conclusions for other amounts of labeled data are similar. We can observe that semi-supervised PCTs and random forests can take considerably more time to train the model than their supervised counterparts, which is expected since they use more data (i.e., additional unlabeled examples) and they also calculate the heuristic score to determine the best splits across all descriptive and target attributes, as opposed to the supervised algorithms that use only the target attributes. The increased learning time is hence the most pronounced on datasets with many attributes, such as Expr-GO and Enron datasets for the HMLC task and Bibtex and Medical datasets for the MLC task. Note that in some cases the learning times between supervised and semi-supervised algorithms are the same. This is because in such cases the $w = 1$ was chosen, i.e., the semi-supervised model is equal to the supervised one. Note that in Table \ref{table:sslpcts_time} the time used to optimize the $w$ parameter is not included.  

\begin{table}[!bt]
	\centering
	\begin{footnotesize}
		\setlength\tabcolsep{4pt}
		\caption{Training times for \textsc{SL-PCT}s and \textsc{SSL-PCT}s.}
		\label{table:sslpcts_time}
		\begin{tabular}{l | rr|rr}
                \hline
                \multicolumn{5}{c}{\textsc{Hierarchical multi-label classification}} \\
			\hline
                Dataset & SL-PCT & SSL-PCT & CLUS-RF & SSL-RF \\
                \hline
                Danish farms & 0.3 & 3.8 & 1.4 & 1.4 \\
                Slovenian rivers & 0.7 & 1.6 & 14.7 & 14.7 \\
                Enron & 1 & 291.2 & 2.1 & 2.1 \\
                Imclef07A & 0.4 & 20.1 & 2.4 & 2.4 \\
                Imclef07D & 0.4 & 8.4 & 1.8 & 47.1 \\
                Diatoms & 7.2 & 100.6 & 11.8 & 11.8 \\
                Cellcycle-GO & 18.5 & 18.5 & 109.9 & 109.9 \\
                Church-GO & 2.3 & 2.3 & 49.4 & 5 \\
                Derisi-GO & 16.8 & 16.8 & 117.1 & 2500.4 \\
                Eisen-GO & 12.1 & 196.8 & 157.3 & 157.3 \\
                Expr-GO & 120.4 & 1031.6 & 267.3 & 267.3 \\
                Pheno-GO & 1.2 & 1.2 & 4.8 & 4.8 \\
                \multicolumn{1}{r|}{\textbf{Average:}} & \textbf{15.1} & \textbf{141.1} & \textbf{61.7} & \textbf{260.4} \\      
                \hline
                \multicolumn{5}{c}{\textsc{Multi-label classification}} \\
			\hline
                Bibtex & 8.1 & 790.2 & 12.9 & 686.4 \\
                Birds & 2.1 & 16.2 & 4.3 & 4.3 \\
                Corel5k & 16.1 & 16.1 & 99.7 & 1314 \\
                Emotions & 0.4 & 0.4 & 1.5 & 7.7 \\
                Enron & 1.7 & 61.8 & 2.7 & 77.8 \\
                Genbase & 0.1 & 0.5 & 0.4 & 0.4 \\
                Mediana & 0.2 & 0.2 & 1.2 & 1.2 \\
                Medical & 1 & 51.8 & 1.3 & 35.1 \\
                Scene & 1.1 & 84.6 & 2.2 & 2.2 \\
                SIGMEA real & 0.1 & 0.1 & 0.5 & 1 \\
                Slovenian rivers & 0.2 & 0.4 & 1.4 & 3.8 \\
                Yeast & 1 & 12 & 3.4 & 3.4 \\
                \multicolumn{1}{r|}{\textbf{Average:}} & \textbf{2.7} & \textbf{86.2} & \textbf{11.0} & \textbf{178.1} \\        
                \hline
            \end{tabular}
        \\[5pt]
        \caption*{The training times are in seconds, obtained for experiments with 500 labeled examples.}
        \end{footnotesize}
\end{table}

\subsection{The influence of unlabeled data}

\begin{figure}[!bt]
	\subfloat[Multi-label classification\label{deltaMLC}]{%
		\includegraphics[height=0.45\textwidth]{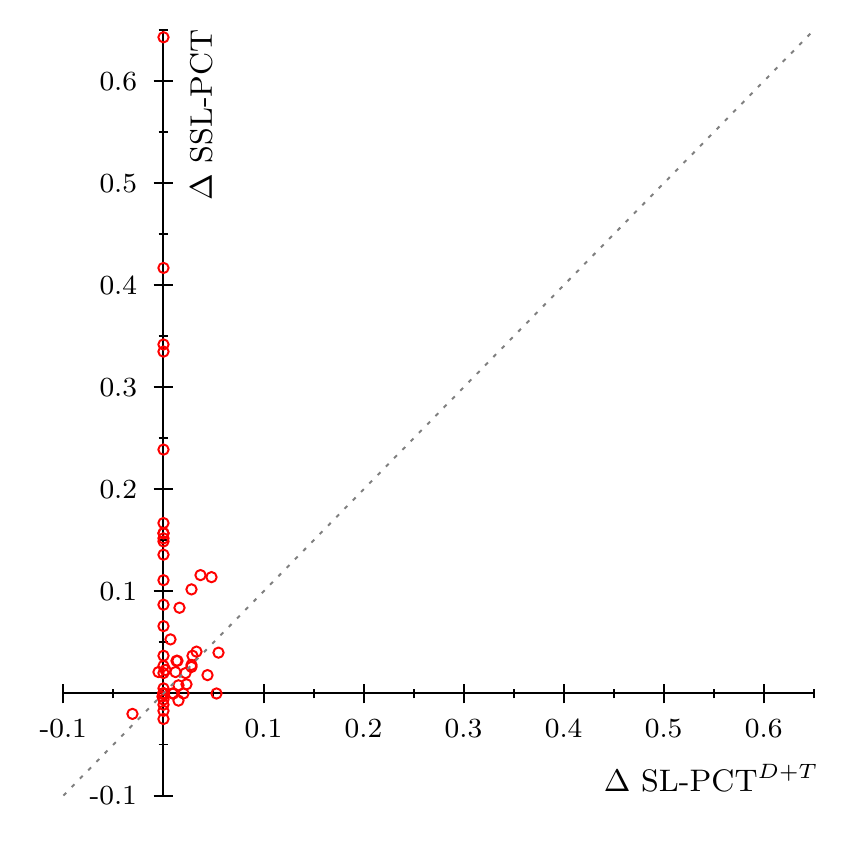}
	}
	\hfill
	\subfloat[Hierarchical multi-label classification\label{deltaHMC}]{%
		\includegraphics[height=0.45\textwidth]{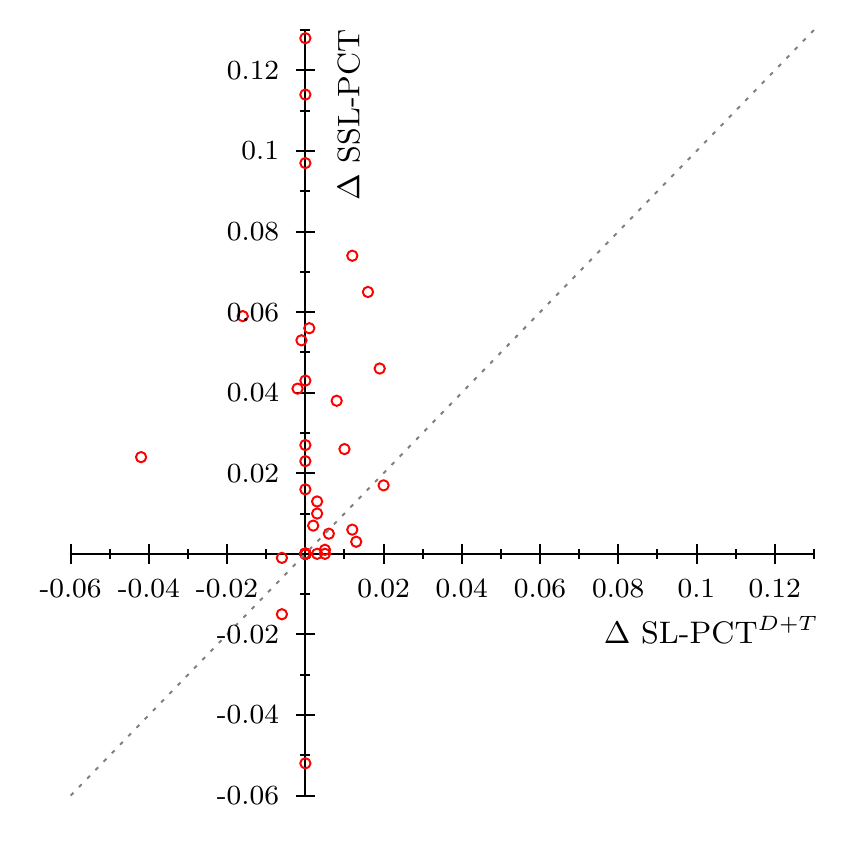}
	}
	\hfill
	\newline
	\caption{\label{fig:sslpcts_sup_vs_ssl} The graph depicts the magnitude of improvement in the predictive performance over supervised PCTs enabled by (i) the variance function that considers both the descriptive and target spaces ($x$-axis) and (ii) unlabeled data and the variance function that considers both the descriptive and target spaces ($y$-axis). This is measured by the difference in the predictive performance of \textsc{SL-PCT}$^{\text{D+T}}$ and SL-PCT ($\Delta\textrm{SL-PCT}^{\text{D+T}}$; $x$-axis), and of SSL-PCT and SL-PCT ($\Delta\textrm{SSL-PCT}$; $y$-axis). The positive values along the $x$ and $y$ axes denote that \textsc{SL-PCT}$^{\text{D+T}}$ or \textsc{SSL-PCT} improve over SL-PCTs, respectively. Clearly, the magnitude of improvement over SL-PCTs along the $y$-axis is much larger than along the $x$-axis, showing that the unlabeled examples are crucial for the performance of the SSL-PCTs. Each dot represents $\mathrm{AU}\overline{\mathrm{PRC}}$ of one experiment (one dataset and one percentage of labeled data; all experiments are considered).}                     
\end{figure}

SSL-PCTs differ from supervised PCTs in two respects: \textit{i)} they use both the descriptive attributes and target variables for the candidate split evaluation, and \textit{ii)} they use unlabeled examples in the training process. We have shown that SSL-PCTs have highly competitive predictive performance with respect to supervised PCTs, but we can still question the source of this improvement. Is this improvement due to the combination of \textit{i)} and \textit{ii)}? Or is \textit{i)} sufficient to yield improvements over supervised PCTs? To answer this question, we compare \textsc{SSL-PCTs} with the supervised modification of PCTs which use both the descriptive attributes and target variables for split evaluation in the same way as SSL-PCTs, but does not use unlabeled data (henceforth, this variant will be denoted as \textsc{SL-PCT}$^{\text{D+T}}$). By using this modification we can evaluate the effect of the unlabeled examples on the predictive performance, since both \textsc{SSL-PCTs} and \textsc{SL-PCT}$^{\text{D+T}}$ are trained using the same algorithms -  the only difference being in the usage of unlabeled data. In these experiments we optimize  parameter $w$ for \textsc{SL-PCT}$^{\text{D+T}}$ via internal 3-fold cross-validation, analogously to \textsc{SSL-PCTs}.

Considering all the datasets and the various percentages of labeled data, the \textsc{SL-PCT}$^{\text{D+T}}$ algorithms perform better than the \textsc{SL-PCT} in 36\% of the cases, the same in 54\% of the cases, and worse in 11\% of the cases. We recall that the corresponding figures for the \textsc{SSL-PCT}s algorithm are 52\%, 39\% and 9\%. Thus, even without the help of unlabeled data, the SSL-PCTs proposed in this work can improve over \textsc{SL-PCT}s, but they have a better chance to do so if they are supplied with unlabeled data. The following result shows that the unlabeled data are indeed the principal component for the success of the \textsc{SSL-PCT}s: The average relative improvement of \textsc{SL-PCT}$^{\text{D+T}}$ over \textsc{SL-PCT} is a mere 4\%, while for \textsc{SSL-PCT} this figure is 40\% (considering only the cases where \textsc{SL-PCT}$^{\text{D+T}}$ and SSL-PCTs improve over SL-PCTs, respectively). This observation, i.e., the importance of unlabeled data, is in line with the findings of \cite{zenko_learning_2007}, where a rule learning process that considers both the descriptive and target spaces is adopted. The results reported in \cite{zenko_learning_2007} show that including the descriptive space in the heuristic was not beneficial for the predictive performance of predictive clustering rules. However, the study was performed in a supervised learning context, i.e., unlabeled examples were not used.

Finally, Figure~\ref{fig:sslpcts_sup_vs_ssl} allows a detailed evaluation of the improvement/degradation of \textsc{SL-PCT}$^{\text{D+T}}$ over \textsc{SL-PCT} and of \textsc{SL-PCT}$^{\text{D+T}}$ over \textsc{SSL-PCT}. As stated previously, the \textsc{SSL-PCT} method outperforms \textsc{SL-PCT} more often than \textsc{SL-PCT}$^{\text{D+T}}$ (this happens when the points are above the diagonal). Furthermore, \textsc{SSL-PCT} yields much larger improvements over \textsc{SL-PCT} than \textsc{SL-PCT}$^{\text{D+T}}$ (for most of the points in the figure, the improvement along the $y$-axis is much larger than the improvement along the $x$-axis). However, there is some complementarity between the two methods. Namely, \textsc{SL-PCT}$^{\text{D+T}}$ sometimes improves over \textsc{SL-PCT} even when this is not the case with \textsc{SSL-PCT} (Figure~\ref{fig:sslpcts_sup_vs_ssl}, the values on the positive side of the x-axis, below the dashed line). 

\section{Conclusions}
\label{sec:conclusions}

In this study, we propose an algorithm for multi-label classification and for hierarchical multi-label classification that works in a semi-supervised learning setting. The method is based on predictive clustering trees and uses both the target and the descriptive space for the evaluation of candidate splits. We executed an extensive empirical study using 24 datasets and we summarize the main findings as follows:
\begin{itemize}
    \item The proposed semi-supervised predictive clustering trees achieve \emph{good predictive performance on both structured output tasks}. On many of the datasets considered, their predictive performance was superior to that of supervised predictive clustering trees.
    \item The \emph{control on the amount of supervision} to be used when learning the proposed semi-supervised predictive clustering trees makes them \emph{safe to use}: They do not degrade the performances with respect to their supervised counterparts, i.e., they either outperform them or have the same performance.
    \item The degree of superiority of semi-supervised over supervised predictive clustering trees does not translate entirely to the tree ensembles, even though semi-supervised random forests often outperform supervised random forests.
    \item Weighting descriptive attributes by their importance may help the predictive performance of semi-supervised predictive clustering in some cases, but the advantages are not great enough to advocate the use of feature weighting by default. Thus, by the principle of Occam's razor, the simpler solution should be preferred, that is, the one without feature weighting.
    \item The semi-supervised trees produce readily interpretable models and are marginally larger than the supervised trees, though the sizes of the trees are reasonable for manual inspection in most cases. Also, this comes with an increase in the computational cost as evidenced by the theoretical and empirical (runtime) analysis of the computational complexity.
\end{itemize}

In future work, we intend to extend the proposed semi-supervised (hierarchical) multi-label classification algorithm to the case where examples are not independent and are accommodated in a network data structure. This would allow us to exploit the semi-supervised learning setting in network data, where the smoothness assumption naturally holds.


\section*{Acknowledgments}
We acknowledge the financial support of the Slovenian Research Agency, via the P2-0103 and J2-2505 grants, and a young researcher grant to the first author, as well as the European Commission, via the H2020 952215 grant TAILOR. The authors would like to thank Lynn Rudd for her help in proofreading the manuscript.

\bibliographystyle{acm}
\bibliography{references.bib}

\begin{thebibliography}{10}

\bibitem{Bauer99:jrnl}
{\sc Bauer, E., and Kohavi, R.}
\newblock An empirical comparison of voting classification algorithms: Bagging,
  boosting, and variants.
\newblock {\em Mach. Learn. 36}, 1 (1999), 105--139.

\bibitem{blockeel_top-down_proc_1998}
{\sc Blockeel, H., De~Raedt, L., and Ramon, J.}
\newblock Top-down induction of clustering trees.
\newblock In {\em Proceeding of the {15th International Conference on Machine
  learning}}. Morgan Kaufmann, San Francisco, CA, 1998, pp.~55--63.

\bibitem{boutell_learning_2004}
{\sc Boutell, M.~R., Luo, J., Shen, X., and Brown, C.~M.}
\newblock Learning multi-label scene classification.
\newblock {\em Pattern Recognit. 37}, 9 (2004), 1757--1771.

\bibitem{Breiman01a:jrnl}
{\sc Breiman, L.}
\newblock Random forests.
\newblock {\em Machine Learning 45}, 1 (2001), 5--32.

\bibitem{Breiman84:book}
{\sc Breiman, L., Friedman, J., Olshen, R., and Stone, C.~J.}
\newblock {\em {Classification and Regression Trees}}.
\newblock Wadsworth \& Brooks, Monterey, CA., 1984.

\bibitem{briggs20139th}
{\sc Briggs, F., Huang, Y., Raich, R., Eftaxias, K., Lei, Z., Cukierski, W.,
  Hadley, S.~F., Hadley, A., Betts, M., Fern, X.~Z., Irvine, J., Neal, L.,
  Thomas, A., Fodor, G., Tsoumakas, G., Ng, H.~W., Nguyen, T. N.~T., Huttunen,
  H., Ruusuvuori, P., Manninen, T., Diment, A., Virtanen, T., Marzat, J.,
  Defretin, J., Callender, D., Hurlburt, C., Larrey, K., and Milakov, M.}
\newblock The 9th annual {MLSP} competition: New methods for acoustic
  classification of multiple simultaneous bird species in a noisy environment.
\newblock In {\em Proceedings of the {IEEE International Workshop on Machine
  Learning for Signal Processing}\/} (2013), IEEE, pp.~1--8.

\bibitem{chapelle_semi-supervised_2006}
{\sc Chapelle, O., Sch\"olkopf, B., and Zien, A.}
\newblock {\em {Semi-supervised Learning}}.
\newblock {MIT} Press, 2006.

\bibitem{chawla_learning_2005}
{\sc Chawla, N., and Karakoulas, G.}
\newblock Learning from labeled and unlabeled data: An empirical study across
  techniques and domains.
\newblock {\em Journal of Artificial Intelligence Research 23}, 1 (2005),
  331--366.

\bibitem{Clare03:phd}
{\sc Clare, A.}
\newblock {\em {Machine Learning and Data Mining for Yeast Functional
  Genomics}}.
\newblock PhD thesis, University of Wales Aberystwyth, Aberystwyth, United
  Kingdom, 2003.

\bibitem{cozman_unlabeled_2002}
{\sc Cozman, F., Cohen, I., and Cirelo, M.}
\newblock Unlabeled data can degrade classification performance of generative
  classifiers.
\newblock In {\em Proceedings of the 15th {International Florida Artificial
  Intelligence Research Society Conference}\/} (2002), {AAAI, Palo Alto,
  California}, pp.~327--331.

\bibitem{cunningham_k-nearest_2007}
{\sc Cunningham, P., and Delany, S.~J.}
\newblock k-nearest neighbour classifiers.
\newblock {\em Multiple Classifier Systems 34\/} (2007), 1--17.

\bibitem{demvsar_statistical_2006}
{\sc Demsar, J.}
\newblock Statistical comparisons of classifiers over multiple data sets.
\newblock {\em J. Mach. Learn. Res. 7\/} (2006), 1--30.

\bibitem{Demsar05-SIGMEA:proc}
{\sc Dem\v{s}ar, D., Debeljak, M., Lavigne, C., and D\v{z}eroski, S.}
\newblock Modelling pollen dispersal of genetically modified oilseed rape
  within the field.
\newblock In {\em Proceedings of the {Annual Meeting of the Ecological Society
  of America}\/} (2005), p.~152.

\bibitem{Demsar06:jrnl}
{\sc Dem\v{s}ar, D., D\v{z}eroski, S., Larsen, T., Struyf, J., Axelsen, J.,
  Pedersen, M., and Krogh, P.}
\newblock Using multi-objective classification to model communities of soil
  microarthopods.
\newblock {\em Ecological Modelling 191}, 1 (2006), 131--143.

\bibitem{dimitrovski2012hierarchical}
{\sc Dimitrovski, I., Kocev, D., Loskovska, S., and D\v{z}eroski, S.}
\newblock Hierarchical classification of diatom images using ensembles of
  predictive clustering trees.
\newblock {\em Ecol. Informatics 7}, 1 (2012), 19--29.

\bibitem{dimitrovski2011hierarchical}
{\sc Dimitrovski, I., Kocev, D., Loskovska, S., and Dzeroski, S.}
\newblock Hierarchical annotation of medical images.
\newblock {\em Pattern Recognit. 44}, 10-11 (2011), 2436--2449.

\bibitem{diplaris2005protein}
{\sc Diplaris, S., Tsoumakas, G., Mitkas, P.~A., and Vlahavas, I.~P.}
\newblock Protein classification with multiple algorithms.
\newblock In {\em Advances in Informatics, 10th Panhellenic Conference on
  Informatics, {PCI} 2005, Volos, Greece, November 11-13, 2005, Proceedings\/}
  (2005), P.~Bozanis and E.~N. Houstis, Eds., vol.~3746 of {\em Lecture Notes
  in Computer Science}, Springer, pp.~448--456.

\bibitem{Duygulu2002}
{\sc Duygulu, P., Barnard, K., de~Freitas, J. F.~G., and Forsyth, D.~A.}
\newblock {\em Object Recognition as Machine Translation: Learning a Lexicon
  for a Fixed Image Vocabulary}.
\newblock Springer, Berlin, 2002, pp.~97--112.

\bibitem{Dzeroski00:jrnl}
{\sc Dzeroski, S., Demsar, D., and Grbovic, J.}
\newblock Predicting chemical parameters of river water quality from
  bioindicator data.
\newblock {\em Appl. Intell. 13}, 1 (2000), 7--17.

\bibitem{elisseeff_kernel_2001}
{\sc Elisseeff, A., and Weston, J.}
\newblock A kernel method for multi-labelled classification.
\newblock In {\em Advances in Neural Information Processing Systems 14 [Neural
  Information Processing Systems: Natural and Synthetic, {NIPS} 2001, December
  3-8, 2001, Vancouver, British Columbia, Canada]\/} (2001), T.~G. Dietterich,
  S.~Becker, and Z.~Ghahramani, Eds., {MIT} Press, pp.~681--687.

\bibitem{guo_extensive_2010}
{\sc Guo, Y., Niu, X., and Zhang, H.}
\newblock An extensive empirical study on semi-supervised learning.
\newblock In {\em {ICDM} 2010, The 10th {IEEE} International Conference on Data
  Mining, Sydney, Australia, 14-17 December 2010\/} (2010), G.~I. Webb, B.~Liu,
  C.~Zhang, D.~Gunopulos, and X.~Wu, Eds., {IEEE} Computer Society,
  pp.~186--195.

\bibitem{DBLP:conf/pkdd/GuoS12}
{\sc Guo, Y., and Schuurmans, D.}
\newblock Semi-supervised multi-label classification - {A} simultaneous
  large-margin, subspace learning approach.
\newblock In {\em Machine Learning and Knowledge Discovery in Databases -
  European Conference, {ECML} {PKDD} 2012, Bristol, UK, September 24-28, 2012.
  Proceedings, Part {II}\/} (2012), P.~A. Flach, T.~D. Bie, and N.~Cristianini,
  Eds., vol.~7524 of {\em Lecture Notes in Computer Science}, Springer,
  pp.~355--370.

\bibitem{DBLP:conf/nips/JeongLKK19}
{\sc Jeong, J., Lee, S., Kim, J., and Kwak, N.}
\newblock Consistency-based semi-supervised learning for object detection.
\newblock In {\em Advances in Neural Information Processing Systems 32: Annual
  Conference on Neural Information Processing Systems 2019, NeurIPS 2019,
  December 8-14, 2019, Vancouver, BC, Canada\/} (2019), H.~M. Wallach,
  H.~Larochelle, A.~Beygelzimer, F.~d'Alch{\'{e}}{-}Buc, E.~B. Fox, and
  R.~Garnett, Eds., pp.~10758--10767.

\bibitem{katakis2008multilabel}
{\sc Katakis, I., Tsoumakas, G., and Vlahavas, I.}
\newblock Multilabel text classification for automated tag suggestion.
\newblock In {\em Proceedings of the ECML/PKDD 2008 Discovery Challenge\/}
  (2008), vol.~75.

\bibitem{Klimt04:proc}
{\sc Klimt, B., and Yang, Y.}
\newblock {\em The {E}nron Corpus: A New Dataset for Email Classification
  Research}, vol.~3201 of {\em Lecture Notes in Computer Science}.
\newblock Springer, Berlin, 2004, pp.~217--226.

\bibitem{Kocev13:jrnl}
{\sc Kocev, D., Vens, C., Struyf, J., and Dzeroski, S.}
\newblock Tree ensembles for predicting structured outputs.
\newblock {\em Pattern Recognit. 46}, 3 (2013), 817--833.

\bibitem{DBLP:journals/tkde/KongNZ13}
{\sc Kong, X., Ng, M.~K., and Zhou, Z.}
\newblock Transductive multilabel learning via label set propagation.
\newblock {\em {IEEE} Trans. Knowl. Data Eng. 25}, 3 (2013), 704--719.

\bibitem{DBLP:journals/kbs/LevaticCKD17}
{\sc Levatic, J., Ceci, M., Kocev, D., and Dzeroski, S.}
\newblock Self-training for multi-target regression with tree ensembles.
\newblock {\em Knowl. Based Syst. 123\/} (2017), 41--60.

\bibitem{levatic2022semi}
{\sc Levati{\'c}, J., Ceci, M., Kocev, D., and D{\v{z}}eroski, S.}
\newblock Semi-supervised predictive clustering trees for (hierarchical)
  multi-label classification.
\newblock {\em arXiv preprint arXiv:2207.09237\/} (2022).

\bibitem{DBLP:journals/eswa/LevaticCSDK20}
{\sc Levatic, J., Ceci, M., Stepisnik, T., Dzeroski, S., and Kocev, D.}
\newblock Semi-supervised regression trees with application to {QSAR}
  modelling.
\newblock {\em Expert Syst. Appl. 158\/} (2020), 113569.

\bibitem{DBLP:journals/isci/LevaticKCD18}
{\sc Levatic, J., Kocev, D., Ceci, M., and Dzeroski, S.}
\newblock Semi-supervised trees for multi-target regression.
\newblock {\em Inf. Sci. 450\/} (2018), 109--127.

\bibitem{DingScott}
{\sc Li, D., and Dick, S.}
\newblock Semi-supervised multi-label classification using an extended
  graph-based manifold regularization.
\newblock {\em Complex {\&} Intelligent Systems 8}, 3 (2022), 1561--1577.

\bibitem{10.1145/3620677}
{\sc Liu, Y., Zhou, X., Kou, H., Zhao, Y., Xu, X., Zhang, X., and Qi, L.}
\newblock Privacy-preserving point-of-interest recommendation based on
  simplified graph convolutional network for geological traveling.
\newblock {\em ACM Trans. Intell. Syst. Technol.\/} (Sep 2023).

\bibitem{10.1145/1835449.1835598}
{\sc Mojdeh, M., and Cormack, G.~V.}
\newblock Semi-supervised spam filtering using aggressive consistency learning.
\newblock In {\em Proceedings of the 33rd International ACM SIGIR Conference on
  Research and Development in Information Retrieval\/} (New York, NY, USA,
  2010), SIGIR '10, Association for Computing Machinery, p.~751–752.

\bibitem{nigam2000text}
{\sc Nigam, K., McCallum, A.~K., Thrun, S., and Mitchell, T.}
\newblock {Text classification from labeled and unlabeled documents using EM}.
\newblock {\em Machine learning 39}, 2-3 (2000), 103--134.

\bibitem{petkovic2022feature}
{\sc Petkovi{\'c}, M., D{\v{z}}eroski, S., and Kocev, D.}
\newblock Feature ranking for semi-supervised learning.
\newblock {\em Machine Learning\/} (2022), 1--30.

\bibitem{PETKOVIC2023101526}
{\sc Petković, M., Levatić, J., Kocev, D., Breskvar, M., and Džeroski, S.}
\newblock Clusplus: A decision tree-based framework for predicting structured
  outputs.
\newblock {\em SoftwareX 24\/} (2023), 101526.

\bibitem{Quinlan93:book}
{\sc Quinlan, R.~J.}
\newblock {\em {{C4.5:} Programs for Machine Learning}}, 1~ed.
\newblock Morgan Kaufmann, 1993.

\bibitem{read2011classifier}
{\sc Read, J., Pfahringer, B., Holmes, G., and Frank, E.}
\newblock Classifier chains for multi-label classification.
\newblock {\em Mach. Learn. 85}, 3 (2011), 333--359.

\bibitem{DBLP:journals/eswa/SantosC14}
{\sc Santos, A.~M., and Canuto, A. M.~P.}
\newblock Applying semi-supervised learning in hierarchical multi-label
  classification.
\newblock {\em Expert Syst. Appl. 41}, 14 (2014), 6075--6085.

\bibitem{DBLP:conf/kdd/ShiSLG20}
{\sc Shi, W., Sheng, V.~S., Li, X., and Gu, B.}
\newblock Semi-supervised multi-label learning from crowds via deep sequential
  generative model.
\newblock In {\em {KDD} '20: The 26th {ACM} {SIGKDD} Conference on Knowledge
  Discovery and Data Mining, Virtual Event, CA, USA, August 23-27, 2020\/}
  (2020), R.~Gupta, Y.~Liu, J.~Tang, and B.~A. Prakash, Eds., {ACM},
  pp.~1141--1149.

\bibitem{Skrjanc01:jrnl}
{\sc Skrjanc, M., Grobelnik, M., and Zupanic, D.}
\newblock Insights offered by data-mining when analyzing media space data.
\newblock {\em Informatica (Slovenia) 25}, 3 (2001), 357--363.

\bibitem{DBLP:conf/cikm/SongMZK21}
{\sc Song, Z., Meng, Z., Zhang, Y., and King, I.}
\newblock Semi-supervised multi-label learning for graph-structured data.
\newblock In {\em {CIKM} '21: The 30th {ACM} International Conference on
  Information and Knowledge Management, Virtual Event, Queensland, Australia,
  November 1 - 5, 2021\/} (2021), G.~Demartini, G.~Zuccon, J.~S. Culpepper,
  Z.~Huang, and H.~Tong, Eds., {ACM}, pp.~1723--1733.

\bibitem{trohidis2008multi}
{\sc Trohidis, K., Tsoumakas, G., Kalliris, G., and Vlahavas, I.~P.}
\newblock Multi-label classification of music into emotions.
\newblock In {\em Proceedings of the 9th {International Conference on Music
  Information Retrieval}\/} (2008), vol.~8, {Drexel University, Philadelphia,
  PA}, pp.~325--330.

\bibitem{DBLP:journals/ml/EngelenH20}
{\sc van Engelen, J.~E., and Hoos, H.~H.}
\newblock A survey on semi-supervised learning.
\newblock {\em Mach. Learn. 109}, 2 (2020), 373--440.

\bibitem{Vens08:jrnl}
{\sc Vens, C., Struyf, J., Schietgat, L., Dzeroski, S., and Blockeel, H.}
\newblock Decision trees for hierarchical multi-label classification.
\newblock {\em Mach. Learn. 73}, 2 (2008), 185--214.

\bibitem{DBLP:conf/aaai/WangLQS020}
{\sc Wang, L., Liu, Y., Qin, C., Sun, G., and Fu, Y.}
\newblock Dual relation semi-supervised multi-label learning.
\newblock In {\em The Thirty-Fourth {AAAI} Conference on Artificial
  Intelligence, {AAAI} 2020, The Thirty-Second Innovative Applications of
  Artificial Intelligence Conference, {IAAI} 2020, The Tenth {AAAI} Symposium
  on Educational Advances in Artificial Intelligence, {EAAI} 2020, New York,
  NY, USA, February 7-12, 2020\/} (2020), {AAAI} Press, pp.~6227--6234.

\bibitem{wilcoxon1945individual}
{\sc Wilcoxon, F.}
\newblock Individual comparisons by ranking methods.
\newblock {\em Biometrics Bulletin 1\/} (1945), 80--83.

\bibitem{witten2005data}
{\sc Witten, I.~H., and Frank, E.}
\newblock {\em {Data Mining: Practical Machine Learning Tools and Techniques}}.
\newblock Morgan Kaufmann, 2005.

\bibitem{YU2010433}
{\sc Yu, D., Varadarajan, B., Deng, L., and Acero, A.}
\newblock Active learning and semi-supervised learning for speech recognition:
  A unified framework using the global entropy reduction maximization
  criterion.
\newblock {\em Computer Speech \& Language 24}, 3 (2010), 433--444.

\bibitem{DBLP:conf/ijcai/ZhaoG15}
{\sc Zhao, F., and Guo, Y.}
\newblock Semi-supervised multi-label learning with incomplete labels.
\newblock In {\em Proceedings of the Twenty-Fourth International Joint
  Conference on Artificial Intelligence, {IJCAI} 2015, Buenos Aires, Argentina,
  July 25-31, 2015\/} (2015), Q.~Yang and M.~J. Wooldridge, Eds., {AAAI} Press,
  pp.~4062--4068.

\bibitem{zhou_semisupervised_2007}
{\sc Zhou, Z., and Li, M.}
\newblock Semisupervised regression with cotraining-style algorithms.
\newblock {\em {IEEE} Trans. Knowl. Data Eng. 19}, 11 (2007), 1479--1493.

\bibitem{zenko_learning_2007}
{\sc Ženko, B.}
\newblock {\em {Learning Predictive Clustering Rules}}.
\newblock {Ph.D. Thesis}, Faculty of Computer Science, University of Ljubljana,
  2007.

\end{thebibliography}

\end{document}